\documentclass{article}

\usepackage[main, final]{neurips_2026}

\usepackage[utf8]{inputenc} 
\usepackage[T1]{fontenc}    
\usepackage[bookmarks=false]{hyperref}       
\usepackage{url}            
\usepackage{booktabs}       
\usepackage{amsfonts}       
\usepackage{amsmath}
\usepackage{nicefrac}       
\usepackage{microtype}      
\usepackage{xcolor}         
\usepackage{hyperref}
\usepackage[capitalise]{cleveref}
\usepackage{graphicx}
\usepackage{algorithm}
\usepackage{algpseudocode}
\usepackage{multirow}
\usepackage[most]{tcolorbox}
\usepackage{wrapfig}
\usepackage{xcolor}
\usepackage{listings}
\usepackage{caption}

\definecolor{denim}{rgb}{0.08, 0.38, 0.74}
\usepackage{enumitem}
\hypersetup{
  colorlinks   = true, 
  urlcolor     = magenta, 
  linkcolor    = denim, 
  citecolor   =  denim,
}

\newcommand{\method}{AVSD}
\newcommand{\Method}{\textbf{\underline{A}}daptive-\textbf{\underline{V}}iew \textbf{\underline{S}}elf-\textbf{\underline{D}}istillation}

\definecolor{darkgreen}{RGB}{0, 100, 0}
\definecolor{salmon}{RGB}{250,128,114}

\title{AVSD: Adaptive-View Self-Distillation by Balancing Consensus and Teacher-Specific Privileged Signals}

\author{%
\textbf{Duy Nguyen}$^{1}$ \quad
\textbf{Hanqi Xiao}$^{1}$ \quad
\textbf{Archiki Prasad}$^{1}$ \quad
\textbf{Zaid Khan}$^{1}$ \quad 
\textbf{Anirban Das}$^{2}$ \quad \\
\textbf{Shi-Xiong Zhang}$^{2}$ \quad
\textbf{Sambit Sahu}$^{2}$ \quad
\textbf{Hyunji Lee}$^{1}$ \quad
\textbf{Elias Stengel-Eskin}$^{3}$ \quad
\textbf{Mohit Bansal}$^{1}$ \\[0.6em]
{\normalfont
$^{1}$UNC Chapel Hill
\qquad
$^{2}$Capital One
\qquad
$^{3}$The University of Texas at Austin
} \\
}

\begin{document}

\maketitle

\begin{abstract}
Self-distillation enables language models to learn on-policy from their own trajectories by using the same model as both student and teacher, with the teacher being conditioned on privileged information unavailable to the student. Such information can come in different types or views, such as solutions, demonstrations, feedback, or final answers. This setup provides dense token-level feedback without relying on a separate external model, but creates a fundamental \textit{asymmetry}: the teacher may rely on view-specific information that the student cannot access at inference time. Moreover, the best type of privileged information is often task-dependent, making it difficult to choose a single teacher view. In this work, we address both these challenges jointly by introducing \method{} (\Method), a novel method of self-distillation with multiple privileged-information views, which reconstructs token-level supervision by separating stable cross-view consensus from view-specific residual signals. 
\method{} identifies the consensus signal shared across views, which provides a reliable update direction, and then selectively adds the view-specific residual signal to adjust the update magnitude when it both aligns with the consensus direction and remains proportionate to the consensus signal. Experiments on math competition benchmarks (AIME24, AIME25, and HMMT25) show that \method{} consistently outperforms both single-view self-distillation baselines and GRPO, achieving average Avg@8 gains of 3.1\% and 2.2\% over the strongest baselines on Qwen3-8B and Qwen3-4B, respectively. Moreover, on code-generation benchmarks (Codeforces, LiveCodeBench v6) using Qwen3-8B, \method{} outperforms the single-view self-distillation baseline by 2.4\% on average.\footnote{Code: \url{https://github.com/duykhuongnguyen/AVSD}.}
\end{abstract}

\section{Introduction}
\begin{figure*}
    \centering
    \includegraphics[width=0.95\linewidth]{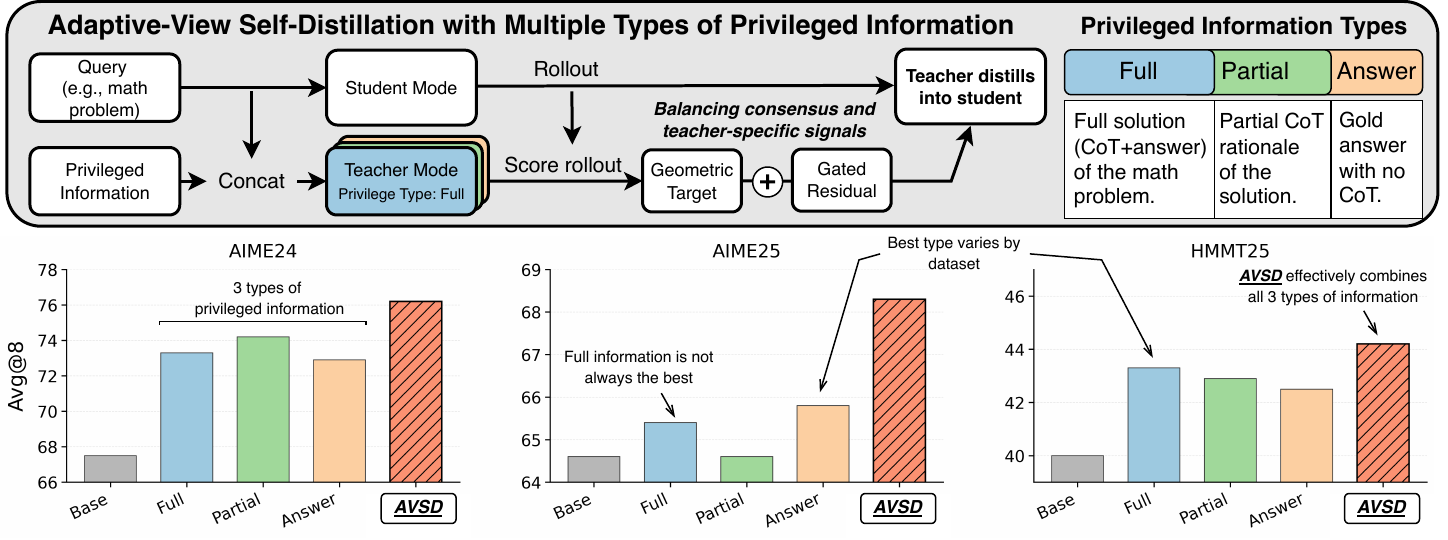}
    \caption{
    Self-distillation leverages access to privileged ground truth information as a teacher to distill into the student. However, performance across datasets on Qwen3-4B varies substantially depending on the type of privileged ground truth information the teacher has access to (full solution, partial reasoning, or only final answer). We show that no single type is uniformly optimal, and that \method{} consistently achieves the best performance across datasets by effectively combining multiple types of privileged information.}
    \label{fig:different_info}
\end{figure*}
Training large language models (LLMs) to solve difficult reasoning tasks, such as competition-level math problems, requires converting available training supervision into useful learning signals. Reinforcement learning with verifiable rewards (RLVR)~\citep{shao2024deepseekmath, deepseekai2025deepseekr1} has become the dominant paradigm for LLM post-training in verifiable tasks such as math and code, but its supervision is typically sparse and outcome level, leading to expensive sampling and limited learning signal when the model fails to produce a successful final solution~\citep{tao2025hybrid}. Distillation offers a complementary source of dense token-level guidance~\citep{hinton2015distilling}, but standard offline distillation suffers from train-test mismatch because the student is trained on trajectories outside its own distribution~\citep{xu2024survey, agarwal2024onpolicy}. On-policy distillation addresses this off-policy issue by training on student-generated rollouts while using teacher probabilities to provide local learning signals~\citep{agarwal2024onpolicy, gu2024minillm}. However, these methods still rely on access to a more capable teacher model, motivating self-distillation methods, which remove the need for an external teacher by employing the student model as its own teacher, supplying a learning signal by giving the teacher privileged information that the student cannot access.
For example, the student model might learn from a teacher with access to different augmented context views, such as solutions, demonstrations, or feedback~\citep{zhao2026self,shenfeld2026self}. 

Current self-distillation methods share a strong and often implicit assumption: once a privileged context type is chosen, it remains fixed throughout training~\citep{shenfeld2026self, penaloza2026privileged}. However, in practice, a variety of privileged information types are possible, with no clear way of determining \emph{a priori} which will be optimal. For example, for a math problem, one may have access to a verified full solution, a final answer, a partial rationale, or a concise hint (see~\cref{fig:different_info} bottom-center, showing differences when using full vs partial vs answer only privileged information). Each type comes with its own limitations, e.g., providing a full solution with chain of thought may lock the teacher into a preexisting thought pattern that does not align with the student's, while providing only the final answer may be more flexible but less informative. These different types of information can result in meaningful performance differences: \citet{penaloza2026privileged} show that different forms of privileged information vary in information density, task utility, and induced student-teacher distribution shift, and that the best choice depends on the task and model regime. ~\citet{yang2026self} show that distribution matching under information asymmetry  contains an irreducible mutual-information gap, so a student trained to imitate a privileged teacher is pressured to encode view-specific correlations it cannot observe at test time.~\citet{kim2026does} show that the most informative privileged form discards important signals that help models identify reasoning mistakes, and degrading out-of-domain mathematical reasoning, even when training traces are correct. These observations suggest that choosing a single privileged view has important consequences for performance and generalization, and that no single view is guaranteed to provide the best learning signals across tasks. This raises the central question of our work: \textbf{can we leverage multiple privileged views simultaneously to construct better token-level on-policy learning signals for self-distillation than any single privileged teacher? }

In this work, we propose \Method{} (\method{}), an on-policy
self-distillation framework that uses multiple types of privileged information rather than committing
to a single teacher context.  
The key intuition is: (1) if different privileged views induce similar token-level updates, then the signal is likely stable and task-relevant. (2) In contrast, if a token is strongly favored by only one view, that signal may still be useful because it may capture complementary information not emphasized by other views, but it is riskier because it may depend on information that the student cannot access at inference time. AVSD formalizes this intuition by separating the teacher signal of multiple views into a shared part and an extra view-specific part. We refer to the shared part as the \emph{consensus} signal: it captures tokens that are consistently supported across privileged views and provides a reliable update direction. We refer to the extra view-specific part as the \emph{residual}: it captures additional support that appears when one or a few views strongly favor a token beyond what is shared by all views. To expose these two parts, AVSD uses two pooled targets: the geometric consensus target and the arithmetic marginal target. The \emph{geometric consensus target} emphasizes tokens that are supported across views, while the \emph{arithmetic marginal target} preserves tokens that receive strong support from at least one view. Rather than distilling from either target directly, AVSD starts from the consensus signal and uses a gate to add the residual only when the different views agree on the promote-or-suppress direction, and the residual magnitude remains proportionate to the consensus magnitude. This allows AVSD to benefit from complementary information in different privileged views while preventing any single view from dominating the consensus token-level learning signal.

We validate \method{} on math and code reasoning benchmarks. Empirically, using three automatically-constructed privileged views for each math problem (full solution, partial solution, and final answer), we show that \method{} outperforms the standard single-view self-distillation baseline~\citep{zhao2026self, shenfeld2026self} and GRPO~\citep{shao2024deepseekmath} on math competition benchmarks (AIME24~\citep{aime24}, AIME25~\citep{aime25}, and HMMT25~\citep{hmmt25}). On Qwen3-4B~\citep{yang2025qwen3}, \method{} improves Avg@8 by 5.5\% over the base model and 2.2\% over the strongest baseline. The gains are consistent at larger scale: on Qwen3-8B, \method{} achieves the best average score, improving over the strongest baseline by 3.1\%. We also observe similar improvements on DeepSeek-R1-Distill-Qwen-7B~\citep{deepseekai2025deepseekr1}, where \method{} improves over the self-distillation baseline by 2.3\% on average. Additionally, on coding benchmarks (Codeforces~\citep{penedo2025codeforces}, LiveCodeBench v6~\citep{jain2025livecodebench}), \method{} improves the single-view self-distillation baseline by 2.4\% on Qwen3-8B. 
Ablations show that consensus-only and arithmetic-only variants both underperform \method{}, and our token-level analysis further confirms that \method{} provides a better learning signal by jointly preserving cross-view agreement and selectively adding gated residual support. 

\section{\method: \Method}
In this section, we present AVSD (Adaptive-View Self-Distillation). In~\cref{sec:problem-setup}, we first formulate on-policy self-distillation through the token-level reverse-KL advantage, then extend it to the multi-view setting where each privileged view induces a different teacher distribution.

At a high level, \method{} builds a teacher signal by separating consensus from view-specific signal. If several views agree that a token should be promoted or suppressed, this provides a reliable anchor for the student update. In~\cref{sec:method_targets}, we capture this shared signal with a \emph{geometric consensus target}, which emphasizes tokens supported across views. In contrast, a token favored by only one or a few views may capture complementary task-relevant information, but may also reflect privileged artifacts. We expose this extra signal with an \emph{arithmetic marginal target}, and define the \emph{residual} as its difference from the geometric consensus target. Finally, in~\cref{sec:method_gate}, we propose a gate to reconstruct the teacher signal by selectively adding this residual when the views agree on the update direction and the residual magnitude remains proportionate to the consensus magnitude.

\subsection{Problem Setup} \label{sec:problem-setup}
Let $\mathcal D$ be a training set. For notation simplicity, we write $(x,r)\sim\mathcal D$ for a generic example, where $x$ is a prompt and $r$ is the training-time privileged information.  The student observes only $x$, while the teacher conditions on $r$ during training. During training, we sample an on-policy rollout
\[
    y=(y_1,\ldots,y_T) \sim P_\theta^S(\cdot \mid x).
\]
We denote the prefix at position $t$ as $h_t = (x,y_{<t})$. The student next-token distribution over vocabulary $\mathcal V$ is
\[
     p_t(v) := P_\theta^S(Y_t=v \mid h_t),
    \:\: v\in\mathcal V,
\]
where $v$ denotes a token type in the vocabulary. Standard single-view self-distillation chooses one view $r$ and uses the same model conditioned on that view as the teacher.  At the same prefix, denote
\[
    q_t(v) :=
    \operatorname{sg}\!\left[
        P_\theta^T(Y_t=v\mid h_t,r)
    \right],
    \:\: v\in\mathcal V,
\]
where $P_\theta^T$ denotes the same model evaluated in the privileged teacher
context and $\operatorname{sg}[\cdot]$ is the stop gradient operator. The single-view reverse-KL~\citep{gu2024minillm, agarwal2024onpolicy} objective is
\[
\mathcal L_{\mathrm{sv}}(\theta)
    =
    \mathbb E_{(x,r)\sim\mathcal D}
    \mathbb E_{y\sim P_\theta^S(\cdot\mid x)}
    \left[
        \frac{1}{|y|}
        \sum_{t=1}^{|y|}
        D_{\mathrm{KL}}\!\left(p_t\,\middle\|\,q_t\right)
    \right].
\]
The negative local gradient of the reverse-KL term can be written
in policy-gradient form as
\[
    -\nabla_\theta
    D_{\mathrm{KL}}\!\left(p_t\,\middle\|\,q_t\right)
    =
    \mathbb E_{v\sim p_t}
    \left[
        A_t(v)\nabla_\theta \log p_t(v)
    \right],  
\]
where
\begin{equation} \label{eq:advantage}
    A_t(v) := \log q_t(v)-\log p_t(v).
\end{equation}
$A_t(v)$ in Eq.~\eqref{eq:advantage} is the token-level distillation advantage.  If $A_t(v)>0$, the teacher assigns token type $v$ higher probability than the student, so gradient descent on the reverse-KL objective promotes $v$; if $A_t(v)<0$, the update suppresses $v$. The full derivation of this policy-gradient form from the reverse-KL objective is given in~\cref{app:rkld_advantage}. We use $A_t(v)$ to describe the sign and relative magnitude of the token-level updates. In the single-view setting, this corresponds to choosing one privileged information type throughout training, such as always conditioning the teacher on a full solution in~\cref{fig:different_info}.

In the multi-view setting, we construct $M$ privileged views $\mathcal{V}(r) = \{r^{(1)}, \ldots, r^{(M)}\}$, where $r^{(m)} = T_m(r)$ is the transformation that preserves task-relevant information while changing which privileged view is revealed to the teacher. For example, for a math problem, views may be a full solution, final answer, or partial rationale. Each view induces a teacher distribution at the same student prefix, which shows how different privileged information types would promote or suppress the same generated token (examples in~\cref{fig:figure2} and~\cref{fig:heatmap}):
\[
    q_t^{(m)}(v)
    := \operatorname{sg}\!\left[
        P^T_\theta(Y_t=v \mid h_t, r^{(m)})
    \right],
    \qquad m=1,\ldots,M,
\]
The corresponding per-view distillation advantage is $
    \Delta_t^{(m)}(v)
    := \log q_t^{(m)}(v) - \log p_t(v).
$
A key question then becomes: \textbf{Given a teacher family $\{q_t^{(m)}\}_{m=1}^M$, how should we weigh and combine distributions from teachers to determine the token-level signal that the student receives?}

\subsection{Consensus and Residual Signal} \label{sec:method_targets}
\begin{figure*}
    \centering
    \includegraphics[width=1\linewidth]{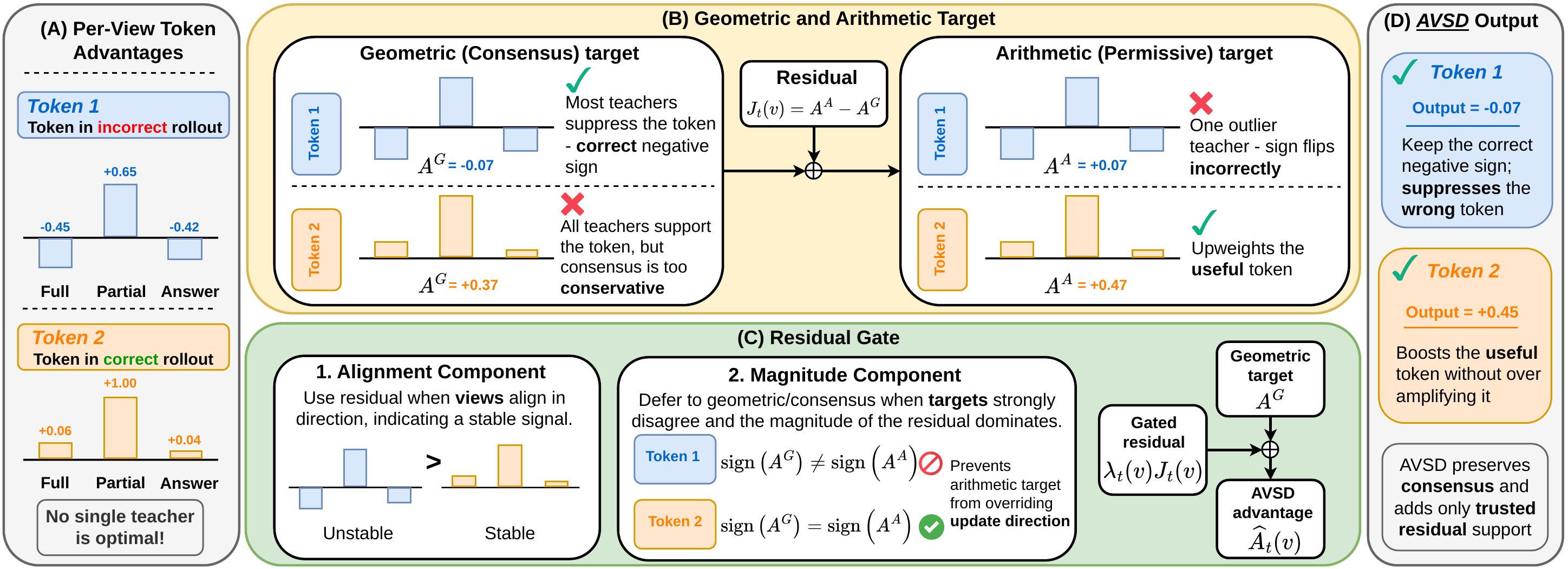}
    \vspace{-1em}
    \caption{(A) Bar graphs show teacher advantages on student tokens. 
    (B) The geometric target demands consensus between teachers (conservative), while the arithmetic target allows individual teachers to contribute more strongly (permissive). The residual is their difference. (C) A two-component gate controls the residual: the alignment component suppresses unstable signals when views conflict, and the magnitude component prevents the residual from flipping the update direction established by consensus. (D) AVSD combines both: it preserves the consensus direction on risky tokens and boosts useful view-specific signals on tokens with high cross-view alignment.
    }
    \label{fig:figure2}
\end{figure*}

Different privileged views can agree or disagree at the same student prefix on the \emph{signed token-level update} induced by each teacher view. For a token $v$, the teacher family provides a stable signal when the per-view advantages $\Delta_t^{(m)}(v)$ have consistent signs and comparable magnitudes: multiple views either agree that $v$ should be promoted or agree that it should be suppressed. This \emph{consensus} is informative because all teachers are conditioned on the same student-visible prefix $h_t$ and differ only in what training-time privileged information they have access to. Therefore, an update that persists across several privileged views is less likely to depend on a particular view-specific artifact and more likely to reflect a \textit{consensus} signal that is compatible with the information available from $h_t$. At the same time, not all useful information needs to be shared by every view: a token may receive strong support from only a subset of views because those views capture reasoning paths complementary to the consensus signal (\cref{fig:figure2} Token 2). However, the same subset-specific support may also reflect privileged artifacts that the student cannot access at test time (\cref{fig:figure2} Token 1) To resolve this, we use two pooled targets: one target captures tokens that are consistently supported across views, while the other captures tokens that are strongly supported by at least one view. Their gap defines the residual signal, which AVSD later gates before adding it to the consensus update.

\paragraph{Geometric Consensus Target.}
The first pooled target captures intersection support: a token receives
large probability only if it is supported across the teacher family. We define
the unnormalized geometric score and its normalized distribution as
\[
    \widetilde q_t^G(v)
    :=
    \left(\prod_{m=1}^M q_t^{(m)}(v)\right)^{1/M}
    =
    \exp\!\left(
        \frac{1}{M}\sum_{m=1}^M \log q_t^{(m)}(v)
    \right),\quad
    q_t^G(v) = \frac{\widetilde q_t^G(v)}
                    {\sum_{u\in\mathcal V}\widetilde q_t^G(u)}.
\]
We call $q_t^G$ the geometric consensus target as it is the normalized geometric mean of the view-conditioned teachers.  The geometric mean downweights tokens that are assigned low probability by any view, and therefore emphasizes tokens that are jointly compatible with the teacher family. Thus, this target can be conservative (\cref{fig:figure2} top-center). Equivalently, $q_t^G$ is the distribution that minimizes average reverse KL to the teacher. The full derivation is in~\cref{app:method_details}.

For token-level updates, we use the unnormalized geometric score, since the normalization contributes only a token-independent constant
\[
    A_t^G(v)
    :=
    \log \widetilde q_t^G(v) - \log p_t(v)
    =
    \frac{1}{M}\sum_{m=1}^M \Delta_t^{(m)}(v).
\]
Thus, $A_t^G(v)$ is the average signed update induced by
the teacher family.

\paragraph{Arithmetic Marginal Target.}
The second pooled target captures union support: a token can receive
high probability if it is strongly supported by any view (\cref{fig:figure2} top-center). We define
\[
    q_t^A(v)
    :=
    \frac{1}{M}\sum_{m=1}^M q_t^{(m)}(v).
\]
This is the arithmetic mean of the view-conditioned teachers.  Equivalently, if
a privileged view index is sampled uniformly at random, $q_t^A$ is the marginal
next-token distribution obtained after integrating out the view.  It is also the
solution to the average forward-KL aggregation problem that finds the single pooled
target that minimizes the average forward-KL from the view-conditioned teachers (see~\cref{sec:app_pooled_target} for details).  
The arithmetic advantage is
\[
    A_t^A(v) := \log q_t^A(v) - \log p_t(v).
\]
The arithmetic target preserves complementary support that appears in only one
or a few views, which can be valuable when different views reveal different
valid reasoning paths.  However, this permissiveness also makes it sensitive to
view-specific artifacts.  For example, a full-solution view may strongly promote
a token because it has access to a reference derivation, while an answer-only
view provides no support for the same token.  Therefore, directly distilling from the
arithmetic target would lead to an update based on a noisy signal.

\paragraph{Cross-View Residual.} Because $q_t^G$ is conservative and $q_t^A$ is permissive, their gap isolates the support retained by the permissive marginal but discarded by the strict consensus:
\[
    A_t^A(v) = A_t^G(v) + J_t(v),
    \qquad
    J_t(v) := \log q_t^A(v) - \log \widetilde q_t^G(v) \geq 0.
\]
We call $J_t(v)$ the \emph{cross-view residual} (\cref{fig:figure2} top-center).  It is non-negative by the arithmetic-geometric mean inequality, since $q_t^A(v)\geq \widetilde q_t^G(v)$ for every token $v$.  Intuitively, $J_t(v)$ measures how much probability mass is present in the arithmetic marginal but absent from strict geometric consensus. A large residual indicates that $q_t^A$ preserves support from only a subset of teachers, while a zero residual indicates exact token-level consensus across all teachers. 
The residual alone does not determine whether the extra support is reliable, because the same large residual can arise from useful complementary evidence or view-specific artifacts. Therefore, we use the consensus as the update direction and use the residual to adjust the magnitude of the update.

\subsection{Gated Reconstruction of Token-Level Supervision via Consensus and Residual Signal} \label{sec:method_gate}
\paragraph{Residual Gate.} The decomposition above suggests a simple principle: always keep view consensus, and add the cross-view residual only when it is reliable. We define the reconstructed advantage:
\[
    \widehat A_t(v)
    :=
    A_t^G(v) + \lambda_t(v) J_t(v),
    \qquad
    \lambda_t(v)\in[0,1].
\]
The gate $\lambda_t(v)$ is designed around two requirements, illustrated by Token 1 and Token 2 in~\cref{fig:figure2}. For Token 1, one view strongly promotes the token, but the other views suppress it. In this case, the arithmetic target would incorrectly flip the update direction, so the gate should defer to the consensus signal. For Token 2, all views promote the token, but one view provides much stronger evidence than the others. In this case, the residual can be useful, because it strengthens an update direction that the views already agree on. However, it should still be bounded so that the extra view-specific signal does not dominate the consensus. These two cases motivate two components of the gate described below.

First, the residual should not be trusted when teacher views heavily contradict each other in terms of update direction. For example, if some views promote a token while others suppress it, the residual is unreliable (\cref{fig:figure2} Alignment Component). We define the alignment component by the ratio between the magnitude of the averaged signed advantage and the average magnitude of the per-view advantages: 
\[
    C_t(v)
    :=
    \frac{|A_t^G(v)|}
    {\frac{1}{M}\sum_{m=1}^M |\Delta_t^{(m)}(v)| + \epsilon} = \frac{\frac{1}{M}|\sum_{m=1}^M \Delta_t^{(m)}(v)|}{\frac{1}{M}\sum_{m=1}^M |\Delta_t^{(m)}(v)| + \epsilon} \in[0,1],
\]
$C_t(v)$ is high when the per-view advantages have the same sign or when unsupported views are nearly neutral, and low when positive and negative view-specific updates cancel. This allows \method{} to distinguish complementary support from conflict-driven support.

Second, the residual should be proportionate to the consensus signal (\cref{fig:figure2} Magnitude Component). If $J_t(v)$ is much larger than $|A_t^G(v)|$, then the arithmetic residual can dominate or even reverse the consensus update. For example, if $A_t^G(v)<0$ but $J_t(v)\gg |A_t^G(v)|$, the arithmetic target would promote a token that the teacher family suppresses on average. To prevent this, we define: 
\[
R_t(v) = \frac{|A_t^G(v)|}{|A_t^G(v)| + J_t(v) + \epsilon}.
\]
The final gate is
\begin{equation} \label{eq:gate}
\lambda_t(v)
=
C_t(v)R_t(v)
=
C_t(v)
\frac{|A_t^G(v)|}
{|A_t^G(v)|+J_t(v)+\epsilon},
\end{equation}
This gate acts as an adaptive regularizer on the residual signal.  It admits residual support when the \emph{consensus is coherent} and the \emph{residual is proportionate}, but suppresses it when the views conflict, when the consensus is weak, or when the residual is too large. Moreover, expanding the gate in Eq.~\eqref{eq:gate} gives $\lambda_t(v)J_t(v)\leq |A_t^G(v)|$, since $C_t(v)\leq 1$ and $J_t(v)\leq |A_t^G(v)|+J_t(v)+\epsilon$ (the full derivation is provided in~\cref{sec:app_gate_proof}); the residual can adjust the update magnitude but cannot reverse the update direction induced by the consensus. Our gate also connects to the failure mode identified by~\citet{yang2026self}: directly matching a teacher conditioned on information unavailable to the student can induce an irreducible mutual-information gap and privileged-information leakage. \method{} addresses this risk in the multi-view setting by preventing subset-supported residuals from freely determining the update direction. \cref{fig:figure2} illustrates how the gate addresses failure modes of the two pooled targets. For Token 1, one view promotes the token while the others suppress it, so \method{} keeps the consensus direction and prevents the residual from flipping the update. For Token 2, the views agree on the positive direction, so the residual can strengthen the update.

\paragraph{Reconstructed Target.} The reconstructed target corresponding to $\widehat A_t$ is obtained by
reweighting the geometric consensus target with the gated residual and
renormalizing:
\[
q_t^\star(v)
=
\frac{
\widetilde q_t^G(v)\exp\!\left(\lambda_t(v)J_t(v)\right)
}{
\sum_{u\in\mathcal V}
\widetilde q_t^G(u)\exp\!\left(\lambda_t(u)J_t(u)\right)
}.
\]
This interpolates tokenwise between two targets. If $\lambda_t(v)=0$, the method reduces to the geometric consensus target. If $\lambda_t(v)=1$ for all $v$, then the method recovers the arithmetic marginal target.

\paragraph{Training Objective.} We train the student by on-policy reverse-KL distillation to the reconstructed target. The distillation loss is
\begin{equation} \label{sec:objective}
\mathcal L_{\mathrm{AVSD}}(\theta)
=
\mathbb E_{(x,r)\sim\mathcal D}
\mathbb E_{y\sim P_\theta^S(\cdot\mid x)}
\left[
\sum_{t=1}^{|y|}
D_{\mathrm{KL}}
\left(
p_\theta(\cdot\mid h_t)
\,\middle\|\,
\operatorname{sg}\!\left[q_t^\star(\cdot)\right]
\right)
\right],
\end{equation}
Gradients flow only through the student distribution $p_\theta(\cdot\mid h_t)$, while the teacher distributions and the reconstructed target are treated as fixed supervision at the current prefix. Equivalently, using the log-derivative form of the reverse-KL objective, the sampled-token update is the reconstructed advantage $\widehat A_t(y_t) = A_t^G(y_t)+\lambda_t(y_t)J_t(y_t).$ Compared with standard single-view self-distillation, \method{} only replaces a single teacher-context evaluation with $M$ view-conditioned teacher evaluations at the same student prefix. It does not require additional student rollouts: all views share the same generated trajectory and can be evaluated in parallel or batched. Since on-policy rollout generation is sequential and typically dominates the training cost, while the additional view-conditioned teacher evaluations are prefill-only forward passes, \method{} only adds modest overhead in practice (details in~\cref{tab:runtime}).  We provide the full training algorithm in Appendix~\ref{sec:app_algo}.

\section{Experiments} \label{sec:exp}
In this section, we evaluate \method{} on math and code reasoning benchmarks across three base models. We first present the experimental setup, including datasets, baselines, and evaluation, and then present the main empirical results.
\subsection{Experimental Setup} \label{sec:exp_setting}
\paragraph{Models.} We evaluate across multiple model families and scales, including Qwen3-4B and Qwen3-8B~\citep{yang2025qwen3}, and DeepSeek-R1-Distill-Qwen-7B~\citep{deepseekai2025deepseekr1}. 

\paragraph{Datasets.} We use the following math and code datasets for training and evaluation:
\begin{itemize}
    \item \textbf{Math datasets:} Following~\citet{zhao2026self}, we use the mathematical reasoning subset of OpenThoughts~\citep{guha2026openthoughts} as the training data. For the multi-view setup, we use three views that are directly available or easily derived from the dataset: the full solution, the final answer, and a partial solution obtained by retaining the initial reasoning steps of the full solution. We evaluate on competition-level mathematics benchmarks including AIME 2024~\citep{aime24}, AIME 2025~\citep{aime25}, and HMMT 2025~\citep{hmmt25}.
    \item \textbf{Code datasets:} We train the models on Codeforces~\citep{penedo2025codeforces}, sampling 5K examples from the Python subset. For each problem, we construct three privileged views: a reference implementation, an algorithmic hint derived from the reference solution, and execution feedback obtained by running the student rollout on available test cases, which includes the generated program and the verifier's pass/fail results.
    For evaluation, we hold out 100 Codeforces problems as an in-domain test set and further evaluate on LiveCodeBench v6~\citep{jain2025livecodebench}.
\end{itemize}
The examples of math and code views are provided in~\cref{sec:app_view_examples}. Similar to~\citep{he2026self}, we report Avg@8 throughout the experiments. We provide the detailed training and evaluation hyperparameters in~\cref{sec:app_exp_setting}.

\paragraph{Baselines.} We compare against three methods trained on the same dataset: (1) SFT, standard supervised fine-tuning, which can be seen as off-policy distillation from a static dataset of the reasoning traces, (2) GRPO~\citep{shao2024deepseekmath}, group relative policy optimization with binary outcome rewards verified against ground-truth answers, (3) OPSD, the single-view on-policy self-distillation baseline~\citep{zhao2026self, shenfeld2026self}. For this baseline, we use the same reverse-KL distillation objective as \method{}, but condition the teacher on a single privileged view: the full solution for math and the reference implementation for code, following the original OPSD setting~\citep{zhao2026self}.

\subsection{Results: \method{} Improves Math and Code Reasoning over GRPO and OPSD}
\begin{table*}
\centering
\small
\caption{Avg@8 accuracy on math and code benchmarks across three base models. The best result within each model block is in \textbf{bold}, and the second best is \underline{underlined}.}
\begin{tabular}{l | c c c| c c| c}
\toprule
\multirow{2}{*}{\textbf{Method}} & \multicolumn{3}{c|}{\textbf{Math}} & \multicolumn{2}{c|}{\textbf{Code}} & \multirow{2}{*}{\textbf{Avg.}} \\
\cmidrule{2-4} \cmidrule{5-6}
 & \textbf{AIME24} & \textbf{AIME25} & \textbf{HMMT25} & \textbf{CF} & \textbf{LCB} & \\
\midrule
\textit{Qwen3-4B} 
& 67.5 & 64.6 & 40.0 & 59.8 & 43.2 & 55.0 \\
\quad + SFT 
& 56.3 & 52.1 & 29.6 & 59.1 & 40.7 & 47.6 \\
\quad + GRPO 
& 70.6 & \underline{65.7} & 41.8 & 61.2 & 43.9 & 56.6 \\
\quad + OPSD 
& \underline{73.3} & 65.4 & \underline{43.3} & \underline{63.6} & \underline{45.2} & \underline{58.2} \\
\quad + \textbf{\method{} (Ours)} 
& \textbf{76.2} & \textbf{68.3} & \textbf{44.2} & \textbf{65.6} & \textbf{45.4} & \textbf{59.9} \\
\midrule
\textit{Qwen3-8B} 
& 71.2 & 65.4 & 41.2 & 60.9 & 47.6 & 57.3 \\
\quad + SFT 
& 70.8 & 63.5 & 41.5 & 61.2 & 43.2 & 56.0 \\
\quad + GRPO 
& 72.5 & \underline{67.1} & \underline{43.0} & 62.8 & 49.9 & 59.1 \\
\quad + OPSD 
& \underline{74.2} & 65.8 & 42.5 & \underline{63.2} & \underline{50.3} & \underline{59.2} \\
\quad + \textbf{\method{} (Ours)} 
& \textbf{75.4} & \textbf{69.6} & \textbf{47.1} & \textbf{65.8} & \textbf{52.4} & \textbf{62.1} \\
\midrule
\textit{DeepSeek-R1-Distill-Qwen-7B} 
& 48.3 & \underline{37.9} & 21.7 & 44.8 & 35.7 & 37.7 \\
\quad + SFT 
& 48.5 & 36.2 & 20.5 & 41.6 & 35.2 & 36.4 \\
\quad + GRPO 
& 51.2 & 37.8 & \underline{23.1} & \underline{46.2} & \underline{38.0} & \underline{39.3} \\
\quad + OPSD 
& \underline{53.3} & 37.5 & 22.6 & 43.6 & 37.6 & 38.9 \\
\quad + \textbf{\method{} (Ours)} 
& \textbf{55.8} & \textbf{39.2} & \textbf{25.4} & \textbf{47.4} & \textbf{39.1} & \textbf{41.4} \\
\bottomrule
\end{tabular}
\label{tab:main_results}
\end{table*}
\cref{tab:main_results} shows that \method{} consistently improves both math and code reasoning across model families. On the math benchmarks, \method{} achieves the best average Avg@8 for all three base models. For Qwen3-4B, \method{} improves over the base model by 5.5\% and outperforms the strongest baseline, OPSD, by 2.2\% on average. The gains are consistent at larger scale: on Qwen3-8B, \method{} improves over the base model by 4.8\% and over the strongest baseline, GRPO, by 3.1\%. We observe similar improvements on DeepSeek-R1-Distill-Qwen-7B, where \method{} improves over the strongest self-distillation baseline by 2.3\% on average.

The benefits also transfer to code generation. On Qwen3-8B, \method{} improves the average code Avg@8 on Codeforces and LiveCodeBench over the base model and OPSD by 4.9\% and 2.4\%, respectively. Similar trends hold for Qwen3-4B and DeepSeek-R1-Distill-Qwen-7B, where \method{} achieves the best code average among the compared methods. Overall, these results demonstrate that \method{} provides consistent gains over GRPO and single-view on-policy self-distillation baselines across both math and code benchmarks.

\section{Analysis}
In this section, we analyze \method{} and examine why it improves over single-view self-distillation. We first ablate the consensus and arithmetic targets to isolate the role of gated reconstruction, and then examine token-level credit allocation to test whether \method{} provides more reliable learning signals on student-generated rollouts. We provide additional results and analyses in~\cref{sec:app_results}.
\begin{figure*}
    \centering
    \includegraphics[width=1.0\linewidth]{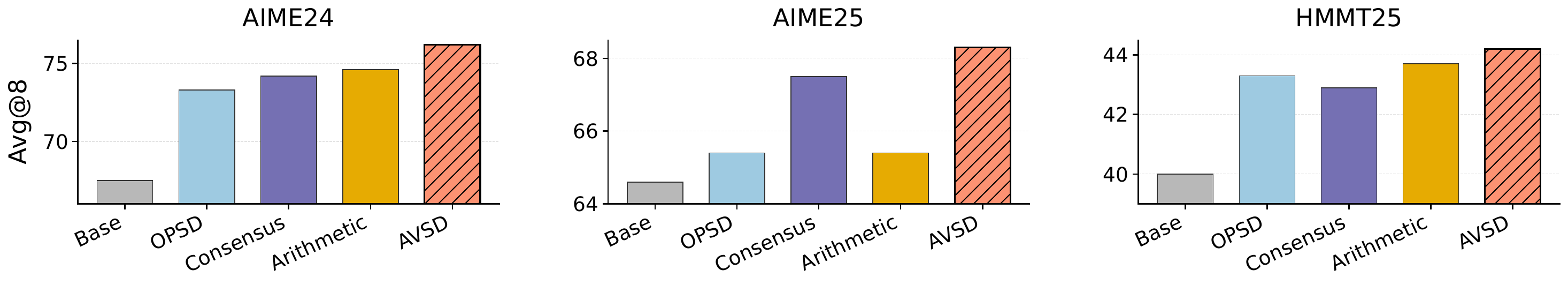}
    \caption{Comparison of single-view OPSD, consensus-only distillation, arithmetic-only distillation, and the full \method{} on Qwen3-4B.~\method{} achieves the best Avg@8 across all datasets.}
    \label{fig:ablation}
\end{figure*}
\subsection{Ablating Consensus and Arithmetic Targets} \label{sec:ablation}
We evaluate the performance of \method{} when using either the consensus target or the arithmetic marginal target alone. The consensus-only variant uses $A_t^G$ as the learning signal and discards the arithmetic-over-consensus residual. The arithmetic-only variant distills directly from $q_t^A$. This comparison isolates the role of our gated reconstruction.~\cref{fig:ablation} shows that neither standalone aggregation is uniformly optimal: the consensus target performs strongly on AIME24 and AIME25 but is slightly worse than OPSD on HMMT25, while the arithmetic target improves on AIME24 and HMMT25 but gives limited gains on AIME25. In contrast, \method{} achieves the best performance across all benchmarks and improves over both standalone multi-view targets. Averaged across the three benchmarks, \method{} outperforms the consensus-only and arithmetic-only variants by 1.4\% and 1.7\%, respectively. This demonstrates that combining the consensus signal with a selectively gated residual is necessary: the consensus target alone can be overly conservative, while the arithmetic target alone can overuse view-specific support. These results support our central claim that reliable multi-view self-distillation requires reconstructing a token-level target from both stable cross-view agreement and carefully filtered residual information.

\subsection{Token-Level Credit Allocation} \label{sec:exp_token_credit}
\begin{wrapfigure}{r}{0.45\textwidth}
    \vspace{-2em}
     \centering
    \includegraphics[width=1.0\linewidth]{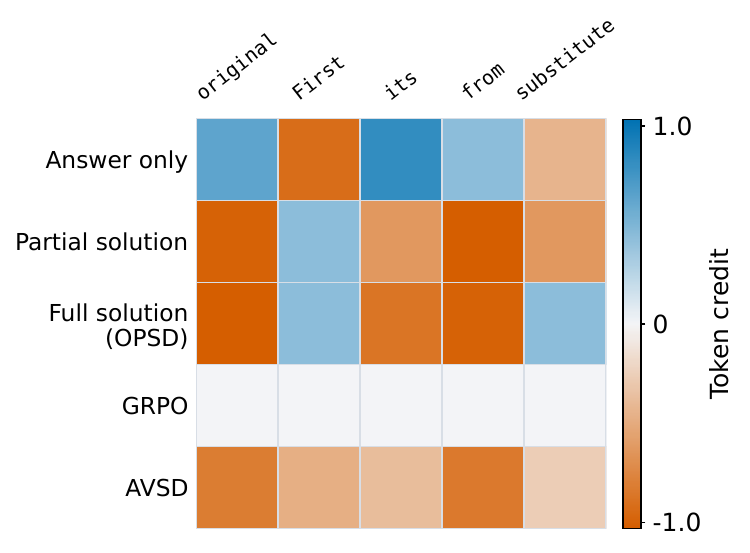}
    \vspace{-1em}
    \caption{
    Example from a single \emph{incorrect rollout}: we show the top-5 generated tokens with largest absolute token-credit magnitude from that rollout. Positive values indicate that the token is encouraged, while negative values indicate that the token is discouraged. For GRPO, the sampled group for the same problem contains no correct rollout, so the outcome-level reward provides no learning signal. \textbf{Single-view self-distillation teachers assign inconsistent or misleading signs across tokens, whereas AVSD provides a more coherent credit signal on key tokens.} Full problem and generation are in~\cref{sec:app_results}.
    }
    \label{fig:heatmap}
\end{wrapfigure}
We further analyze whether the token-level learning signal produced by \method{} assigns credit in the desired direction. The sampled-token advantage determines whether the probability of a generated token is increased or decreased. For an incorrect rollout, generated tokens should generally receive negative credit, indicating that the method suppresses erroneous solutions. To evaluate this behavior, we collect student rollouts whose final answers are incorrect and compute the sampled-token advantage at every generated token. For OPSD and \method{}, we select the top 20 tokens with the largest absolute advantage magnitude, corresponding to the tokens where the distillation signal has the strongest effect on training, and measure the fraction of these tokens whose credit has the incorrect (positive) sign. For GRPO, which does not provide dense token-level supervision, we instead report the fraction of cases with no learning signal (e.g., while rollouts should have either positive or negative signs depending on correctness, by construction, most GRPO methods give no signal when no sampled rollout is correct). 
The heatmap in~\cref{fig:heatmap} visualizes this behavior at the token level for the top 5 tokens in one rollout: each column is a generated token from the incorrect rollout, and each cell shows whether a method increases or decreases that token's probability. Since the rollout is incorrect, the desired behavior is to assign negative credit to the high-impact tokens, i.e., the generated tokens with the largest absolute advantage magnitudes and therefore the strongest effect on the training loss. Single-view baselines produce mixed signs across tokens, suggesting that their feedback depends strongly on the chosen privileged view, whereas \method{} assigns more consistently negative credit to the key tokens while avoiding the near-zero signal of GRPO. Quantitatively, GRPO has no learning signal in 29.6\% of all examples, OPSD assigns incorrect-sign credit to 12.3\% of high-impact tokens, whereas \method{} reduces this rate to only 2.9\% (see~\cref{fig:sign},~\cref{sec:app_results}). These results show that \method{} provides a substantially more reliable token-level learning signal than both sparse-reward GRPO and single-view self-distillation.

\section{Related Work} \label{sec:related_work}
\paragraph{Learning from Privileged Information.}
Learning from privileged information studies how training-time side information unavailable at inference can improve a model~\citep{lopez2015unifying}. In LLM post-training, such information can be reference solutions, final answers, demonstrations, reward models, or environment feedback. Prior work uses these signals in several ways. RLVR methods provide sparse verifiable rewards for math and code~\citep{shao2024deepseekmath,deepseekai2025deepseekr1}, while self-improvement and bootstrapping methods construct additional supervision from model-generated data, verified reasoning traces, or reward models~\citep{zelikman2022star,wang2023selfinstruct, nguyen2025laser}. Other approaches use privileged information to improve exploration or curriculum construction:~\citet{qu2026pope} leverages oracle solution prefixes for on-policy exploration; ~\citet{yan2026learning} learns under off-policy guidance from privileged rollouts; ~\citet{chen2026nudging,liao2026self,wang2026skill} transform privileged trajectories into abstract hints. Synthetic-curriculum methods such as ~\citet{sundaram2026teaching,chen2026cog} generate easier or more informative training instances to bridge low-signal regimes. These methods demonstrate that privileged or derived information can substantially improve learning when direct reward signal is sparse. However, they generally focus on how to obtain better trajectories, hints, or curricula, rather than how to construct a token-level learning signal when multiple privileged views are simultaneously available. Moreover, the choice of privileged format remains consequential: different forms of privileged information vary in information density, task utility, and induced student--teacher distribution shift~\citep{penaloza2026privileged, li2026rethinking}. Our work is complementary to these approaches: rather than proposing a new source of privileged information or exploration, \method{} provides a principled way to aggregate multiple privileged views into a better learning signal by separating cross-view consensus from view-specific residual support.

\paragraph{On-Policy Self-Distillation and Teacher Aggregation.} Self-distillation methods remove the external teacher needed in on-policy distillation by using the same model under richer conditioning.~\citet{zhao2026self} condition on reference math solution,~\citet{shenfeld2026self, ye2026policy} distill demonstration-conditioned behavior, and reinforcement learning via self-distillation uses feedback, successful rollouts, or reward-conditioned self-revisions as privileged context~\citep{hubotter2026reinforcement,song2026expanding,he2026self}.~\citet{penaloza2026privileged} further study teacher-student transfer when the teacher has privileged information and the student must act without it. These methods show that the distillation objectives can serve as a useful token-level advantage. \citet{yang2026self} identify a key failure of direct privileged-teacher matching under information asymmetry and anchors the update direction to verifiable rewards while using self-distillation mainly to adjust token-level update strength. Related hybrid RL-distillation methods combine policy optimization with distillation losses~\citep{ding2026hdpo,li2026unifying, xu2025kdrl}, and ensemble distillation methods aggregate teachers by averaging logits or votes. However, these methods do not characterize which parts of the teacher signal are stable across views versus specific to a privileged context. In contrast, \method{} derives the aggregation structure from the reverse-KL learning signal itself. Thus, our method complements RLSD and other hybrid RL-distillation works: rather than relying on the binary rewards (which are only available in verifiable settings) to determine the update direction for a single privileged teacher in self-distillation, \method{} reconstructs the token-level learning signal from the geometry of a multi-view privileged teacher family.

\section{Conclusion}
In this work, we introduced \method{}, a multi-view framework for privileged self-distillation. Instead of distilling from a single privileged teacher, \method{} treats different views of the same training-time artifact as a teacher family and reconstructs a better token-level learning signal from their shared structure. This allows the student to benefit from complementary privileged information while reducing the influence of conflicting or unreliable view-specific signals. Across math and code reasoning benchmarks and multiple model families, \method{} consistently outperforms single-view self-distillation and GRPO. Our analysis further shows \method{}'s performance generally improves as the number of views increases and it is robust to a single misleading view. that More broadly, our results suggest that moving beyond fixed single-view privileged supervision and explicitly modeling agreement and disagreement across multiple teacher views provides a robust and general approach for constructing dense on-policy learning signals.

\section*{Acknowledgements}
This work was supported by NSF-CAREER Award 1846185, NSF AI Engage Institute DRL2112635, Capital One Research Award, Apple PhD Fellowship, and NDSEG PhD Fellowship. The views contained in this article are those of the authors and not of the funding agency.

\bibliography{bibliography}

@misc{wang2026skill,
      title={Skill-SD: Skill-Conditioned Self-Distillation for Multi-turn LLM Agents}, 
      author={Hao Wang and Guozhi Wang and Han Xiao and Yufeng Zhou and Yue Pan and Jichao Wang and Ke Xu and Yafei Wen and Xiaohu Ruan and Xiaoxin Chen and Honggang Qi},
      year={2026},
      eprint={2604.10674},
      archivePrefix={arXiv},
      primaryClass={cs.LG},
      url={https://arxiv.org/abs/2604.10674}, 
}

@misc{liao2026self,
      title={Self-Hinting Language Models Enhance Reinforcement Learning}, 
      author={Baohao Liao and Hanze Dong and Xinxing Xu and Christof Monz and Jiang Bian},
      year={2026},
      eprint={2602.03143},
      archivePrefix={arXiv},
      primaryClass={cs.LG},
      url={https://arxiv.org/abs/2602.03143}, 
}

@misc{sundaram2026teaching,
      title={Teaching Models to Teach Themselves: Reasoning at the Edge of Learnability}, 
      author={Shobhita Sundaram and John Quan and Ariel Kwiatkowski and Kartik Ahuja and Yann Ollivier and Julia Kempe},
      year={2026},
      eprint={2601.18778},
      archivePrefix={arXiv},
      primaryClass={cs.LG},
      url={https://arxiv.org/abs/2601.18778}, 
}

@inproceedings{wang2023selfinstruct,
    title = "Self-Instruct: Aligning Language Models with Self-Generated Instructions",
    author = "Wang, Yizhong  and
      Kordi, Yeganeh  and
      Mishra, Swaroop  and
      Liu, Alisa  and
      Smith, Noah A.  and
      Khashabi, Daniel  and
      Hajishirzi, Hannaneh",
    booktitle = "Proceedings of the 61st Annual Meeting of the Association for Computational Linguistics (Volume 1: Long Papers)",
    year = "2023",
    url = "https://aclanthology.org/2023.acl-long.754/"
}

@inproceedings{zelikman2022star,
title={{ST}aR: Bootstrapping Reasoning With Reasoning},
author={Eric Zelikman and Yuhuai Wu and Jesse Mu and Noah Goodman},
booktitle={Advances in Neural Information Processing Systems},
editor={Alice H. Oh and Alekh Agarwal and Danielle Belgrave and Kyunghyun Cho},
year={2022},
url={https://openreview.net/forum?id=_3ELRdg2sgI}
}

@misc{lopez2015unifying,
      title={Unifying distillation and privileged information}, 
      author={David Lopez-Paz and Léon Bottou and Bernhard Schölkopf and Vladimir Vapnik},
      year={2016},
      eprint={1511.03643},
      archivePrefix={arXiv},
      primaryClass={stat.ML},
      url={https://arxiv.org/abs/1511.03643}, 
}

@inproceedings{yan2026learning,
title={Learning to Reason under Off-Policy Guidance},
author={Jianhao Yan and Yafu Li and Zican Hu and Zhi Wang and Ganqu Cui and Xiaoye Qu and Yu Cheng and Yue Zhang},
booktitle={The Thirty-ninth Annual Conference on Neural Information Processing Systems},
year={2026},
url={https://openreview.net/forum?id=vO8LLoNWWk}
}

@misc{xu2025kdrl,
      title={KDRL: Post-Training Reasoning LLMs via Unified Knowledge Distillation and Reinforcement Learning}, 
      author={Hongling Xu and Qi Zhu and Heyuan Deng and Jinpeng Li and Lu Hou and Yasheng Wang and Lifeng Shang and Ruifeng Xu and Fei Mi},
      year={2025},
      eprint={2506.02208},
      archivePrefix={arXiv},
      primaryClass={cs.LG},
      url={https://arxiv.org/abs/2506.02208}, 
}

@misc{ding2026hdpo,
      title={HDPO: Hybrid Distillation Policy Optimization via Privileged Self-Distillation}, 
      author={Ken Ding},
      year={2026},
      eprint={2603.23871},
      archivePrefix={arXiv},
      primaryClass={cs.LG},
      url={https://arxiv.org/abs/2603.23871}, 
}

@misc{qu2026pope,
      title={POPE: Learning to Reason on Hard Problems via Privileged On-Policy Exploration}, 
      author={Yuxiao Qu and Amrith Setlur and Virginia Smith and Ruslan Salakhutdinov and Aviral Kumar},
      year={2026},
      eprint={2601.18779},
      archivePrefix={arXiv},
      primaryClass={cs.LG},
      url={https://arxiv.org/abs/2601.18779}, 
}

@misc{shao2024deepseekmath,
      title={DeepSeekMath: Pushing the Limits of Mathematical Reasoning in Open Language Models}, 
      author={Zhihong Shao and Peiyi Wang and Qihao Zhu and Runxin Xu and Junxiao Song and Xiao Bi and Haowei Zhang and Mingchuan Zhang and Y. K. Li and Y. Wu and Daya Guo},
      year={2024},
      eprint={2402.03300},
      archivePrefix={arXiv},
      primaryClass={cs.CL},
      url={https://arxiv.org/abs/2402.03300}, 
}

@inproceedings{agarwal2024onpolicy,
title={On-Policy Distillation of Language Models: Learning from Self-Generated Mistakes},
author={Rishabh Agarwal and Nino Vieillard and Yongchao Zhou and Piotr Stanczyk and Sabela Ramos Garea and Matthieu Geist and Olivier Bachem},
booktitle={The Twelfth International Conference on Learning Representations},
year={2024},
url={https://openreview.net/forum?id=3zKtaqxLhW}
}

@inproceedings{gu2024minillm,
title={Mini{LLM}: Knowledge Distillation of Large Language Models},
author={Yuxian Gu and Li Dong and Furu Wei and Minlie Huang},
booktitle={The Twelfth International Conference on Learning Representations},
year={2024},
url={https://openreview.net/forum?id=5h0qf7IBZZ}
}

@misc{zhao2026self,
      title={Self-Distilled Reasoner: On-Policy Self-Distillation for Large Language Models}, 
      author={Siyan Zhao and Zhihui Xie and Mengchen Liu and Jing Huang and Guan Pang and Feiyu Chen and Aditya Grover},
      year={2026},
      eprint={2601.18734},
      archivePrefix={arXiv},
      primaryClass={cs.LG},
      url={https://arxiv.org/abs/2601.18734}, 
}

@misc{shenfeld2026self,
      title={Self-Distillation Enables Continual Learning}, 
      author={Idan Shenfeld and Mehul Damani and Jonas Hübotter and Pulkit Agrawal},
      year={2026},
      eprint={2601.19897},
      archivePrefix={arXiv},
      primaryClass={cs.LG},
      url={https://arxiv.org/abs/2601.19897}, 
}

@misc{hubotter2026reinforcement,
      title={Reinforcement Learning via Self-Distillation}, 
      author={Jonas Hübotter and Frederike Lübeck and Lejs Behric and Anton Baumann and Marco Bagatella and Daniel Marta and Ido Hakimi and Idan Shenfeld and Thomas Kleine Buening and Carlos Guestrin and Andreas Krause},
      year={2026},
      eprint={2601.20802},
      archivePrefix={arXiv},
      primaryClass={cs.LG},
      url={https://arxiv.org/abs/2601.20802}, 
}

@misc{penaloza2026privileged,
      title={Privileged Information Distillation for Language Models}, 
      author={Emiliano Penaloza and Dheeraj Vattikonda and Nicolas Gontier and Alexandre Lacoste and Laurent Charlin and Massimo Caccia},
      year={2026},
      eprint={2602.04942},
      archivePrefix={arXiv},
      primaryClass={cs.LG},
      url={https://arxiv.org/abs/2602.04942}, 
}

@misc{kim2026does,
      title={Why Does Self-Distillation (Sometimes) Degrade the Reasoning Capability of LLMs?}, 
      author={Jeonghye Kim and Xufang Luo and Minbeom Kim and Sangmook Lee and Dohyung Kim and Jiwon Jeon and Dongsheng Li and Yuqing Yang},
      year={2026},
      eprint={2603.24472},
      archivePrefix={arXiv},
      primaryClass={cs.CL},
      url={https://arxiv.org/abs/2603.24472}, 
}

@misc{yang2026self,
      title={Self-Distilled RLVR}, 
      author={Chenxu Yang and Chuanyu Qin and Qingyi Si and Minghui Chen and Naibin Gu and Dingyu Yao and Zheng Lin and Weiping Wang and Jiaqi Wang and Nan Duan},
      year={2026},
      eprint={2604.03128},
      archivePrefix={arXiv},
      primaryClass={cs.LG},
      url={https://arxiv.org/abs/2604.03128}, 
}

@misc{he2026self,
      title={Self-Distillation Zero: Self-Revision Turns Binary Rewards into Dense Supervision}, 
      author={Yinghui He and Simran Kaur and Adithya Bhaskar and Yongjin Yang and Jiarui Liu and Narutatsu Ri and Liam Fowl and Abhishek Panigrahi and Danqi Chen and Sanjeev Arora},
      year={2026},
      eprint={2604.12002},
      archivePrefix={arXiv},
      primaryClass={cs.CL},
      url={https://arxiv.org/abs/2604.12002}, 
}

@misc{yang2025qwen3,
      title={Qwen3 Technical Report}, 
      author={An Yang and Anfeng Li and Baosong Yang and Beichen Zhang and Binyuan Hui and Bo Zheng and Bowen Yu and Chang Gao and Chengen Huang and Chenxu Lv and Chujie Zheng and Dayiheng Liu and Fan Zhou and Fei Huang and Feng Hu and Hao Ge and Haoran Wei and Huan Lin and Jialong Tang and Jian Yang and Jianhong Tu and Jianwei Zhang and Jianxin Yang and Jiaxi Yang and Jing Zhou and Jingren Zhou and Junyang Lin and Kai Dang and Keqin Bao and Kexin Yang and Le Yu and Lianghao Deng and Mei Li and Mingfeng Xue and Mingze Li and Pei Zhang and Peng Wang and Qin Zhu and Rui Men and Ruize Gao and Shixuan Liu and Shuang Luo and Tianhao Li and Tianyi Tang and Wenbiao Yin and Xingzhang Ren and Xinyu Wang and Xinyu Zhang and Xuancheng Ren and Yang Fan and Yang Su and Yichang Zhang and Yinger Zhang and Yu Wan and Yuqiong Liu and Zekun Wang and Zeyu Cui and Zhenru Zhang and Zhipeng Zhou and Zihan Qiu},
      year={2025},
      eprint={2505.09388},
      archivePrefix={arXiv},
      primaryClass={cs.CL},
      url={https://arxiv.org/abs/2505.09388}, 
}

@inproceedings{guha2026openthoughts,
title={OpenThoughts: Data Recipes for Reasoning Models},
author={Etash Kumar Guha and Ryan Marten and Sedrick Keh and Negin Raoof and Georgios Smyrnis and Hritik Bansal and Marianna Nezhurina and Jean Mercat and Trung Vu and Zayne Rea Sprague and Ashima Suvarna and Benjamin Feuer and Leon Liangyu Chen and Zaid Khan and Eric Frankel and Sachin Grover and Caroline Choi and Niklas Muennighoff and Shiye Su and Wanjia Zhao and John Yang and Shreyas Pimpalgaonkar and Kartik sharma and Charlie Cheng-Jie Ji and Yichuan Deng and Sarah M Pratt and Vivek Ramanujan and Jon Saad-Falcon and Stutee Acharya and Jeffrey Li and Achal Dave and Alon Albalak and Kushal Arora and Blake Wulfe and Chinmay Hegde and Greg Durrett and Sewoong Oh and Mohit Bansal and Saadia Gabriel and Aditya Grover and Kai-Wei Chang and Vaishaal Shankar and Aaron Gokaslan and Mike A Merrill and Tatsunori Hashimoto and Yejin Choi and Jenia Jitsev and Reinhard Heckel and Maheswaran Sathiamoorthy and Alex Dimakis and Ludwig Schmidt},
booktitle={The Fourteenth International Conference on Learning Representations},
year={2026},
url={https://openreview.net/forum?id=7xjoTuaNmN}
}

@misc{li2026unifying,
      title={Unifying Group-Relative and Self-Distillation Policy Optimization via Sample Routing}, 
      author={Gengsheng Li and Tianyu Yang and Junfeng Fang and Mingyang Song and Mao Zheng and Haiyun Guo and Dan Zhang and Jinqiao Wang and Tat-Seng Chua},
      year={2026},
      eprint={2604.02288},
      archivePrefix={arXiv},
      primaryClass={cs.LG},
      url={https://arxiv.org/abs/2604.02288}, 
}

@misc{deepseekai2025deepseekr1,
      title={DeepSeek-R1: Incentivizing Reasoning Capability in LLMs via Reinforcement Learning}, 
      author={DeepSeek-AI},
      year={2025},
      eprint={2501.12948},
      archivePrefix={arXiv},
      primaryClass={cs.CL},
      url={https://arxiv.org/abs/2501.12948}, 
}

@misc{aime24,
      title={American Invitational Mathematics Examination (AIME) 2024}, 
      author={Zhang, Yifan and Math-AI, Team},
      year={2024},
}

@misc{aime25,
      title={American Invitational Mathematics Examination (AIME) 2025}, 
      author={Zhang, Yifan and Math-AI, Team},
      year={2025},
}

@misc{hmmt25,
  title = {MathArena: Evaluating LLMs on Uncontaminated Math Competitions},
  author = {Mislav Balunovic and Jasper Dekoninck and Ivo Petrov and Nikola Jovanovic and Martin Vechev},
  copyright = {MIT},
  url = {https://matharena.ai/},
  publisher = {SRI Lab, ETH Zurich},
  month = feb,
  year = {2025},
}

@inproceedings{kwon2023efficient,
  title={Efficient Memory Management for Large Language Model Serving with PagedAttention},
  author={Woosuk Kwon and Zhuohan Li and Siyuan Zhuang and Ying Sheng and Lianmin Zheng and Cody Hao Yu and Joseph E. Gonzalez and Hao Zhang and Ion Stoica},
  booktitle={Proceedings of the ACM SIGOPS 29th Symposium on Operating Systems Principles},
  year={2023}
}

@misc{penedo2025codeforces,
      title={CodeForces}, 
      author={Guilherme Penedo and Anton Lozhkov and Hynek Kydlícek and Loubna Ben Allal and Edward Beeching and Agustín Piqueres Lajarín and Quentin Gallouédec and Nathan Habib and Lewis Tunstall and Leandro von Werra},
      year={2025},
      publisher = {Hugging Face},
      journal = {Hugging Face repository},
      howpublished = {\url{https://huggingface.co/datasets/open-r1/codeforces}}
}

@inproceedings{jain2025livecodebench,
title={LiveCodeBench: Holistic and Contamination Free Evaluation of Large Language Models for Code},
author={Naman Jain and King Han and Alex Gu and Wen-Ding Li and Fanjia Yan and Tianjun Zhang and Sida Wang and Armando Solar-Lezama and Koushik Sen and Ion Stoica},
booktitle={The Thirteenth International Conference on Learning Representations},
year={2025},
url={https://openreview.net/forum?id=chfJJYC3iL}
}

@misc{ye2026policy,
      title={On-Policy Context Distillation for Language Models}, 
      author={Tianzhu Ye and Li Dong and Xun Wu and Shaohan Huang and Furu Wei},
      year={2026},
      eprint={2602.12275},
      archivePrefix={arXiv},
      primaryClass={cs.CL},
      url={https://arxiv.org/abs/2602.12275}, 
}

@misc{li2026rethinking,
      title={Rethinking On-Policy Distillation of Large Language Models: Phenomenology, Mechanism, and Recipe}, 
      author={Yaxuan Li and Yuxin Zuo and Bingxiang He and Jinqian Zhang and Chaojun Xiao and Cheng Qian and Tianyu Yu and Huan-ang Gao and Wenkai Yang and Zhiyuan Liu and Ning Ding},
      year={2026},
      eprint={2604.13016},
      archivePrefix={arXiv},
      primaryClass={cs.LG},
      url={https://arxiv.org/abs/2604.13016}, 
}

@misc{song2026expanding,
      title={Expanding the Capabilities of Reinforcement Learning via Text Feedback}, 
      author={Yuda Song and Lili Chen and Fahim Tajwar and Remi Munos and Deepak Pathak and J. Andrew Bagnell and Aarti Singh and Andrea Zanette},
      year={2026},
      eprint={2602.02482},
      archivePrefix={arXiv},
      primaryClass={cs.LG},
      url={https://arxiv.org/abs/2602.02482}, 
}

@inproceedings{chen2026nudging,
title={Nudging the Boundaries of {LLM} Reasoning},
author={Justin Chen and Xiangyu Peng and Prafulla Kumar Choubey and Kung-Hsiang Huang and Jiaxin Zhang and Mohit Bansal and Chien-Sheng Wu},
booktitle={The Fourteenth International Conference on Learning Representations},
year={2026},
url={https://openreview.net/forum?id=hfNnQHkTtv}
}

@misc{chen2026cog,
      title={Cog-DRIFT: Exploration on Adaptively Reformulated Instances Enables Learning from Hard Reasoning Problems}, 
      author={Justin Chih-Yao Chen and Archiki Prasad and Zaid Khan and Joykirat Singh and Runchu Tian and Elias Stengel-Eskin and Mohit Bansal},
      year={2026},
      eprint={2604.04767},
      archivePrefix={arXiv},
      primaryClass={cs.LG},
      url={https://arxiv.org/abs/2604.04767}, 
}

@misc{tao2025hybrid,
      title={Hybrid Reinforcement: When Reward Is Sparse, It's Better to Be Dense}, 
      author={Leitian Tao and Ilia Kulikov and Swarnadeep Saha and Tianlu Wang and Jing Xu and Sharon Li and Jason E Weston and Ping Yu},
      year={2025},
      eprint={2510.07242},
      archivePrefix={arXiv},
      primaryClass={cs.CL},
      url={https://arxiv.org/abs/2510.07242}, 
}

@misc{xu2024survey,
      title={A Survey on Knowledge Distillation of Large Language Models}, 
      author={Xiaohan Xu and Ming Li and Chongyang Tao and Tao Shen and Reynold Cheng and Jinyang Li and Can Xu and Dacheng Tao and Tianyi Zhou},
      year={2024},
      eprint={2402.13116},
      archivePrefix={arXiv},
      primaryClass={cs.CL},
      url={https://arxiv.org/abs/2402.13116}, 
}

@misc{hinton2015distilling,
      title={Distilling the Knowledge in a Neural Network}, 
      author={Geoffrey Hinton and Oriol Vinyals and Jeff Dean},
      year={2015},
      eprint={1503.02531},
      archivePrefix={arXiv},
      primaryClass={stat.ML},
      url={https://arxiv.org/abs/1503.02531}, 
}

@article{abbas2009kullback,
  title={A Kullback-Leibler view of linear and log-linear pools},
  author={Abbas, Ali E},
  journal={Decision Analysis},
  volume={6},
  number={1},
  pages={25--37},
  year={2009},
  publisher={INFORMS}
}

@inproceedings{nguyen2025laser,
title={{LAS}eR: Learning to Adaptively Select Reward Models with Multi-Arm Bandits},
author={Duy Nguyen and Archiki Prasad and Elias Stengel-Eskin and Mohit Bansal},
booktitle={The Thirty-ninth Annual Conference on Neural Information Processing Systems},
year={2025},
url={https://openreview.net/forum?id=tSpWkTFASC}
}
\bibliographystyle{plainnat}

\appendix
\crefalias{section}{appendix}
\crefalias{subsection}{appendix} 

\newpage

\section{Limitations and Broader Impact} \label{sec:app_limitations}
\paragraph{Limitations.} Our work focuses primarily on reasoning tasks such as math and code. Although the formulation of \method{} is general, validating it in broader settings, such as tool-use agents, is a valuable direction for future work. The method also depends on the quality of the constructed views, as well as the underlying data quality. For example, in math reasoning, noisy annotated solutions may lead to less reliable teacher signals. In addition, \method{} is instantiated with reverse-KL on-policy self-distillation. While the reconstructed target can in principle be used with other distribution-matching losses, our advantage interpretation and bounded-residual guarantee require further study for other objectives, as discussed in~\cref{sec:app_other_loss}.

\paragraph{Broader Impact.} This work aims to improve how language models learn from information that is available during training, such as solutions, demonstrations, or feedback. By making better use of this information, \method{} may help improve model performance on reasoning tasks such as math and code. Better models could support education, coding assistance, scientific problem solving, and other applications. However, stronger reasoning models can also be misused, and improvements in post-training methods should be accompanied by careful evaluation and deployment safeguards. In addition, training-time information may contain biases and artifacts from the data collection process. Although \method{} is designed to reduce reliance on any single privileged source, it does not remove the need for high-quality data and evaluation for unintended behavior. 

\section{Additional Details for \method{}} \label{app:method_details}
In this section, we provide additional details for our method. In~\cref{app:rkld_advantage}, we derive the reverse-KL sampled-token advantage used in on-policy self-distillation. In~\cref{sec:app_pooled_target}, we derive the arithmetic and geometric pooled targets and show how they correspond to natural aggregation objectives over the teacher family. In~\cref{sec:app_gate_proof}, we provide further details on the gated reconstruction mechanism and the residual-boundedness property used in the main method. In~\cref{sec:app_algo}, we provide the detailed training algorithm for \method{}. In~\cref{sec:app_other_loss}, we discuss the applicability to other distillation losses.

\subsection{Reverse-KL Token Advantage}
\label{app:rkld_advantage}

Fix a student prefix $h_t$ and a stop-gradient teacher distribution $q_t$.  Let
$p_t(v)=p_\theta(Y_t=v\mid h_t)$.  The local reverse-KL loss is
\[
    \ell_t(\theta)
    =
    D_{\mathrm{KL}}(p_t\|q_t)
    =
    \sum_{v\in\mathcal V} p_t(v)
    \left(\log p_t(v)-\log q_t(v)\right).
\]
Taking the gradient with respect to the student parameters,
\[
\begin{aligned}
    \nabla_\theta \ell_t
    &=
    \sum_v \nabla_\theta p_t(v)
    \left(\log p_t(v)-\log q_t(v)+1\right) \\
    &=
    \mathbb E_{v\sim p_t}
    \left[
        \left(\log p_t(v)-\log q_t(v)+1\right)
        \nabla_\theta \log p_t(v)
    \right].
\end{aligned}
\]
The constant $1$ is a baseline because
$\mathbb E_{v\sim p_t}[\nabla_\theta \log p_t(v)]=0$.  Therefore,
\[
    \nabla_\theta \ell_t
    =
    \mathbb E_{v\sim p_t}
    \left[
        \left(\log p_t(v)-\log q_t(v)\right)
        \nabla_\theta \log p_t(v)
    \right].
\]
Equivalently, minimizing reverse KL is a policy-gradient update with
distillation advantage
\[
    A_t(v) = \log q_t(v)-\log p_t(v),
\]
since gradient descent on $\ell_t$ moves in the direction
\[
    \mathbb E_{v\sim p_t}
    \left[
        A_t(v)\nabla_\theta \log p_t(v)
    \right].
\]
This is the advantage used in the main text to interpret whether a token type is
promoted or suppressed.  The sampled-token estimator replaces the full
expectation over $v\sim p_t$ by the generated token $y_t$ at the visited prefix.

\subsection{Derivations of Two Pooled Targets} \label{sec:app_pooled_target}
Fix a prefix $h_t$ and write $q_m = q_t^{(m)}$.

\paragraph{Geometric Target.} The geometric target is
\[
q_t^G = \arg \min_q \frac{1}{M} \sum_{m=1}^M D_{\text{KL}}(q \parallel q_m).
\]
Expanding gives
\[
\frac{1}{M} \sum_m D_{\text{KL}}(q \parallel q_m) = \sum_v q(v) \log q(v) - \sum_v q(v) \frac{1}{M} \sum_m \log q_m(v).
\]
Let
\[
\tilde{q}_G(v) = \exp \left( \frac{1}{M} \sum_m \log q_m(v) \right).
\]
The Lagrangian is
\[
\mathcal{L}(q, \nu) = \sum_v q(v) \log q(v) - \sum_v q(v) \log \tilde{q}_G(v) + \nu \left( \sum_v q(v) - 1 \right).
\]

Taking derivatives gives
\[
\log q(v) + 1 - \log \tilde{q}_G(v) + \nu = 0,
\]
so
\[
q(v) \propto \tilde{q}_G(v).
\]
Therefore,
\[
q_t^G(v) = \frac{\tilde{q}_t^G(v)}{\sum_u \tilde{q}_t^G(u)}.
\]

\paragraph{Arithmetic Target.} The arithmetic target is
\[
q_t^A = \arg \min_q \frac{1}{M} \sum_{m=1}^M D_{\text{KL}}(q_m \parallel q).
\]

Expanding the terms that depend on $q$,
\[
\frac{1}{M} \sum_m D_{\text{KL}}(q_m \parallel q) = - \sum_v \left( \frac{1}{M} \sum_m q_m(v) \right) \log q(v) + \text{const.}
\]

The minimizer of cross-entropy with target distribution
\[
\frac{1}{M} \sum_m q_m
\]
is exactly
\[
q_t^A(v) = \frac{1}{M} \sum_m q_t^{(m)}(v).
\]
The geometric and arithmetic targets also connect to classical opinion pooling: linear pools correspond to arithmetic aggregation, while log-linear pools and products of experts combine distributions through normalized products~\citep{abbas2009kullback}. The same aggregation perspective naturally extends to non-uniform view weights. If each view is assigned a prefix-level reliability weight $w_t^{(m)} \ge 0$ with $\sum_m w_t^{(m)}=1$, then the geometric and arithmetic pooled targets become the corresponding weighted KL barycenters by replacing the uniform factor $1/M$ with $w_t^{(m)}$ in the objectives above. This yields the same interpretation: the geometric target captures weighted cross-view consensus, while the arithmetic target captures weighted view-marginal support. In our main experiments, we use uniform weights to keep the method parameter-free and avoid introducing additional view-reliability hyperparameters.

\begin{algorithm}[t]
\caption{Multi-View Self-Distillation with Gated Reconstruction. In our experiments, we find that $M=3$ works well.}
\label{alg:mvsd}
\begin{algorithmic}[1]
\Require Student model \(p_\theta\), dataset \(\mathcal D=\{(x,r)\}\), view transformations
\(\{T_m\}_{m=1}^M\), small constant \(\epsilon>0\).
\For{each training iteration}
    \State Sample a minibatch \(\{(x_i,r_i)\}\sim\mathcal D\).
    \For{each example \((x,r)\)}
        \State Construct privileged views \(r^{(m)}=T_m(r)\) for \(m=1,\ldots,M\).
        \State Sample an on-policy rollout \(y\sim p_\theta(\cdot\mid x)\).
        \For{each prefix \(h_t=(x,y_{<t})\)}
            \State Compute student distribution \(p_t(\cdot)=p_\theta(\cdot\mid h_t)\).
            \For{each view \(m=1,\ldots,M\)}
                \State Compute teacher distribution
                \[
                q_t^{(m)}(\cdot)=\operatorname{sg}\left[p_\theta(\cdot\mid h_t,r^{(m)})\right].
                \]
            \EndFor
            \State Compute
            \[
            q_t^A(v)=\frac1M\sum_m q_t^{(m)}(v),
            \qquad
            \widetilde q_t^G(v)=
            \left(\prod_m q_t^{(m)}(v)\right)^{1/M}.
            \]
            \State Compute residual
            \[
            J_t(v)=\log q_t^A(v)-\log \widetilde q_t^G(v).
            \]
            \State Compute per-view advantages
            \[
            \Delta_t^{(m)}(v)=\log q_t^{(m)}(v)-\log p_t(v),
            \qquad
            A_t^G(v)=\frac1M\sum_m\Delta_t^{(m)}(v).
            \]
            \State Compute gate
            \[
            C_t(v)=
            \frac{|A_t^G(v)|}
            {\frac1M\sum_m|\Delta_t^{(m)}(v)|+\epsilon},
            \qquad
            \lambda_t(v)=
            C_t(v)
            \frac{|A_t^G(v)|}{|A_t^G(v)|+J_t(v)+\epsilon}.
            \]
            \State Reconstruct teacher
            \[
            q_t^\star(v)
            =
            \frac{
            \widetilde q_t^G(v)\exp(\lambda_t(v)J_t(v))
            }{
            \sum_{u\in\mathcal V}
            \widetilde q_t^G(u)\exp(\lambda_t(u)J_t(u))
            }.
            \]
            \State Accumulate loss
            \[
            \mathcal L_{\mathrm{AVSD}}
            \leftarrow
            \mathcal L_{\mathrm{AVSD}}
            +
            D_{\mathrm{KL}}
            \left(
            p_t(\cdot)
            \,\middle\|\,
            \operatorname{sg}\!\left[q_t^\star(\cdot)\right]
            \right).
            \]
        \EndFor
    \EndFor
    \State Update $\theta$ by minimizing $\mathcal L_{\mathrm{AVSD}}(\theta)$.
\EndFor
\end{algorithmic}
\end{algorithm}

\subsection{Bounded Residual Property} \label{sec:app_gate_proof}
For the gate
\[
\lambda_t(v)=C_t(v)
\frac{|A_t^G(v)|}{|A_t^G(v)|+J_t(v)+\epsilon},
\]
where \(C_t(v)\in[0,1]\), we have
\[
0\le \lambda_t(v)J_t(v)
=
C_t(v)
\frac{|A_t^G(v)|J_t(v)}
{|A_t^G(v)|+J_t(v)+\epsilon}
\le |A_t^G(v)|.
\]
Therefore, the reconstructed advantage
\[
\widehat A_t(v)=A_t^G(v)+\lambda_t(v)J_t(v)
\]
preserves the direction of the consensus update: if \(A_t^G(v)>0\), then
\(\widehat A_t(v)\ge 0\); if \(A_t^G(v)<0\), then
\(\widehat A_t(v)\le 0\). Thus, the residual can strengthen a positive
consensus update or soften a negative consensus update, but cannot flip the
consensus-induced direction.

\begin{table}[t]
\centering
\small
\caption{Runtime overhead of \method{} compared with single-view self-distillation. We report average wall-clock training time per gradient update and relative overhead using Qwen3-4B under the same batch size and hardware setup. \method{} uses the same student rollouts as OPSD and only adds view-conditioned teacher evaluations.}
\label{tab:runtime_overhead}
\begin{tabular}{lccc}
\toprule
\textbf{Method}  & \textbf{Time / step (s)}  & \textbf{Relative time} \\
\midrule
OPSD &  41.5 & 1.00$\times$ \\
\method{} & 46.7  & 1.13$\times$ \\
\bottomrule
\label{tab:runtime}
\end{tabular}
\end{table}
\subsection{Detailed Training Algorithm} \label{sec:app_algo}
Algorithm~\ref{alg:mvsd} summarizes the full training procedure. The method preserves the on-policy nature
of self-distillation: all teacher distributions are evaluated on prefixes generated by the current student. Compared to single-view self-distillation, the only additional cost is evaluating multiple privileged views for the same prefix. In practice, these view-conditioned teacher forward passes can be batched. We report the run time comparison with single-view OPSD in~\cref{tab:runtime}. \method{} increases training time only modestly, from 41.5s to 46.7s per step, corresponding to a 1.13$\times$ relative cost, since all views share the same student rollout and only add batched teacher evaluations.

\subsection{Applicability to other Distillation Losses} \label{sec:app_other_loss}
Although we derive \method{} with reverse KL, the teacher reconstruction itself is not tied to a particular divergence. The multi-view aggregation step constructs a reconstructed target $q_t^\star$ from the teacher family $\{q_t^{(m)}\}_{m=1}^M$, and this target can in principle be used with other distribution-matching losses, such as forward KL,
\[
D_{\mathrm{KL}}\!\left(q_t^\star \,\|\, p_\theta(\cdot\mid h_t)\right),
\]
or symmetric objectives such as Jensen--Shannon divergence. However, the policy-gradient interpretation in our method is specific to the reverse-KL objective: the sampled-token update is driven by the reconstructed advantage
\[
\widehat A_t(y_t)=A_t^G(y_t)+\lambda_t(y_t)J_t(y_t),
\]
which is naturally evaluated on student-generated tokens and therefore remains on-policy. In contrast, forward KL places expectation under the reconstructed teacher $q_t^\star$, requiring full-vocabulary supervision or teacher-sampled tokens, and does not yield the same sampled-token advantage interpretation. As a result, while the reconstructed teacher $q_t^\star$ can be paired with other distillation losses, our main design and guarantees are most directly aligned with reverse-KL on-policy self-distillation. Exploring whether the same multi-view reconstruction improves forward-KL or symmetric distillation objectives is an interesting direction for future work.

\section{Implementation Details of Experiments} \label{sec:app_exp_setting}
\paragraph{Training and Evaluation Configuration.} We provide the detailed training hyperparameters for different methods in~\cref{tab:training_hyperparameters}. Following prior work~\citep{zhao2026self, li2026unifying}, we disable the Qwen3 thinking mode during training. Prior work has also observed that Qwen models with thinking mode enabled tend to produce long, exploratory chains of thought, which can be harder to learn from~\citep{kim2026does}. Therefore, we leave this as an open problem for future work.

For evaluation, we use vLLM~\citep{kwon2023efficient} with the following settings for all models: temperature $=0.6$, top-$p=0.95$, top-$k=20$, min-$p=0$, and max tokens $38912$.

\begin{table}[t]
\centering
\caption{Training hyperparameters for GRPO and self-distillation methods.}
\label{tab:training_hyperparameters}
\begin{tabular}{llcc}
\toprule
\textbf{Category} & \textbf{Parameter} & \textbf{GRPO} & \textbf{OPSD/AVSD} \\
\midrule
\multirow{5}{*}{Training}
& Learning Rate & $5 \times 10^{-6}$ & $5 \times 10^{-6}$ \\
& Effective Batch Size & 16 & 16 \\
& Optimizer & AdamW & AdamW \\
& Gradient Clip Norm & 0.1 & 0.1 \\
& Training Steps & 500 & 200 \\
\midrule
\multirow{3}{*}{LoRA}
& LoRA Rank ($r$) & 64 & 64 \\
& LoRA Alpha ($\alpha$) & 128 & 128 \\
& LoRA Target Modules &
\multicolumn{2}{c}{\begin{tabular}{c}
q\_proj, k\_proj, v\_proj, o\_proj, \\
gate\_proj, up\_proj, down\_proj
\end{tabular}} \\
\midrule
\multirow{3}{*}{Generation}
& Number of Rollouts & 8 & 1 \\
& Sampling Temperature & 1.0 & 0.7 \\
& Inference Engine & vLLM & vLLM \\
\bottomrule
\end{tabular}
\end{table}

\paragraph{GPUs.} Most of the experiments are run on 4 A6000 GPUs. Some experiments such as training for Qwen3-8B and GRPO baselines are run on 4 A100 GPUs.

\section{Additional Experiments} \label{sec:app_results}
\subsection{Privileged Information Analysis}

\begin{figure}[h]
    \centering
    \includegraphics[width=1.0\linewidth]{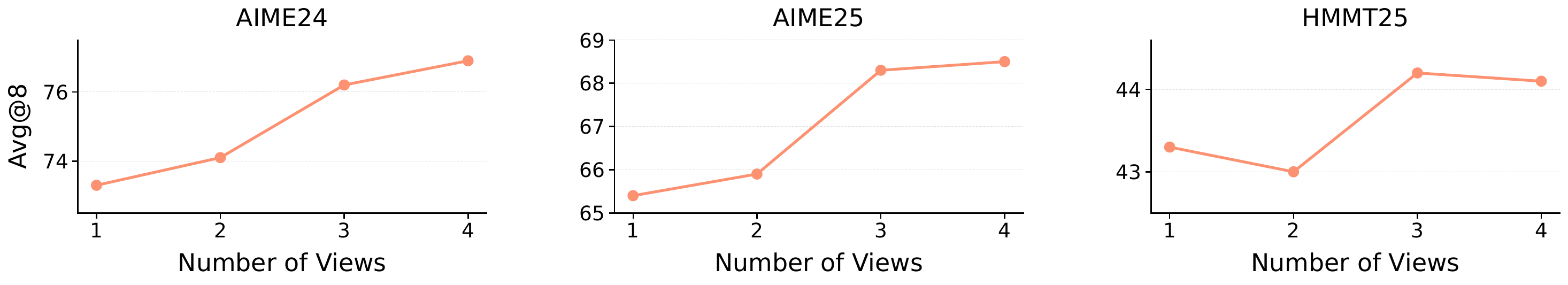}
    \vspace{-1em}
    \caption{Scaling the number of privileged math views for \method{} on Qwen3-4B.}
    \label{fig:num_views}
\end{figure}

\begin{figure}[h]
    \centering
    \includegraphics[width=0.7\linewidth]{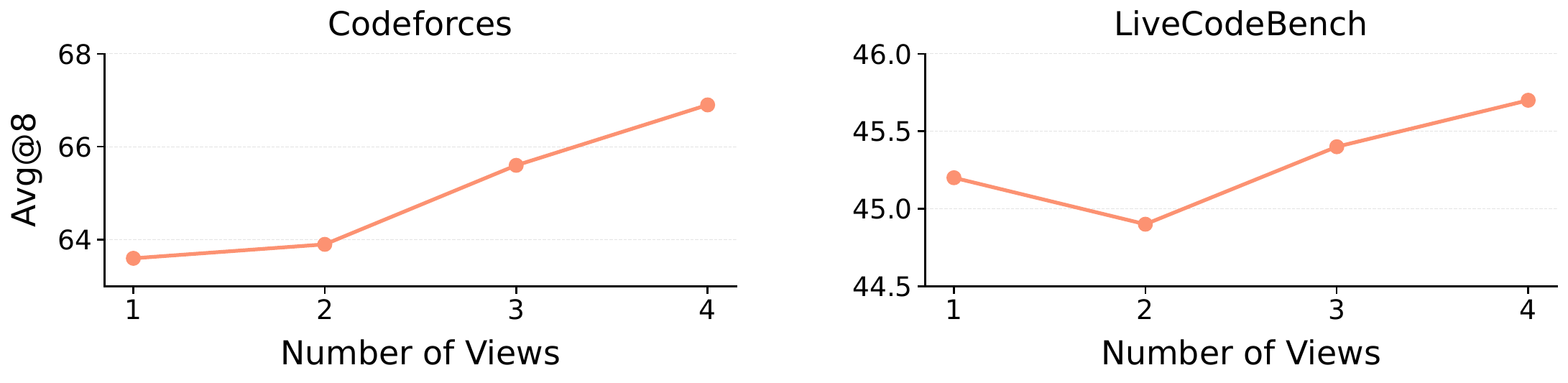}
    \caption{Scaling the number of privileged code views for \method{} on Qwen3-4B. Final performance generally improves as more complementary privileged views are added.}
    \label{fig:num_views_code}
\end{figure}

\paragraph{Scaling the Number of Views.} We study how performance changes as we increase the number of privileged views used by \method{}. For math, we start from a single full-solution view and incrementally add a partial-solution view, the final-answer view used in our main experiments, and a fourth view that conditions the teacher on the model's own attempt together with the reference solution. As shown in \cref{fig:num_views}, performance generally improves as the view family becomes more informative and diverse. On Qwen3-4B, moving from one to three math views improves Avg@8 by 2.9\% on AIME24 and 2.8\% on AIME25. Adding a fourth math view provides only a small additional gain on AIME24 and AIME25 (+0.7\% and +0.2\%), while remaining comparable on HMMT25, suggesting diminishing returns once the main sources of privileged information are covered.

We observe a similar trend on code, where we start from a single reference-implementation view and incrementally add complementary code views.~\cref{fig:num_views_code} shows that moving from one to three views improves Codeforces by 2.0\%. Unlike the math setting, the fourth code view continues to provide a clearer benefit, with gains of 3.3\% and 0.5\% over the single-view setting, respectively. Overall, these results suggest that \method{} benefits from richer teacher families across both math and code: multiple views expose more stable cross-view signal for gated reconstruction, while additional views can still help when they provide complementary information.

\paragraph{Effect of Privileged-View Choice.} In~\cref{fig:different_info}, we show that for math reasoning no single privileged view is uniformly optimal across benchmarks. We further examine whether the same phenomenon holds in the code domain on Qwen3-8B. As shown in~\cref{tab:code-results}, different code views lead to different strengths: the reference-implementation view performs best among single-view OPSD variants on Codeforces, while the feedback view performs best on LiveCodeBench and achieves the strongest single-view average. This confirms that the optimal privileged view is task-dependent in code as well. In contrast, \method{} combines all views and achieves the best performance on both benchmarks, improving over the strongest single-view OPSD baseline by 2.6\% on Codeforces and 1.5\% on LiveCodeBench. These results further support our motivation that multi-view self-distillation can construct a stronger and more robust learning signal than committing to any single privileged view.
\begin{table}[t]
\centering
\small
\caption{
Effect of privileged-view choice in code self-distillation. We report Avg@8 on Codeforces and LiveCodeBench with Qwen3-8B.
}
\label{tab:code-results}
\begin{tabular}{llccc}
\toprule
\textbf{Method} 
& \textbf{View(s)}
& \textbf{Codeforces}
& \textbf{LiveCodeBench}
& \textbf{Avg.} \\
\midrule
Base 
&  
& 60.9 
& 47.6 
& 54.3 \\
\midrule
OPSD 
& code 
& \underline{63.2} 
& 50.3 
& 56.8 \\
OPSD 
& hint 
& 61.3 
& 48.2 
& 54.8 \\
OPSD 
& feedback 
& 62.8 
& \underline{50.9} 
& \underline{56.9} \\
\midrule
\textbf{\method{} (Ours)}
& all 
& \textbf{65.8} 
& \textbf{52.4} 
& \textbf{59.1} \\
\bottomrule
\end{tabular}
\end{table}

\paragraph{Robustness to Misleading and Highly Correlated Views.} We next test whether \method{} remains robust when the teacher family contains either unreliable or redundant privileged information under two different settings. For the \emph{misleading-view} setting, we corrupt the answer-only view by replacing its correct final answer with a randomly sampled, different final answer from another training example, while leaving the full- and partial-solution views unchanged. For the \emph{highly correlated-view} setting, we replace the complementary partial-solution and answer-only views with prefixes retaining 90\% and 95\% of the full solution, yielding three closely related views containing 90\%, 95\%, and 100\% of the same solution.

\cref{tab:view_robustness} shows that \method{} is robust to misleading views. Corrupting the answer-only view reduces the average score by only 0.4\% and leaves HMMT25 performance unchanged. This suggests that when one view conflicts with the remaining informative views, the residual gate can limit the influence of unreliable view-specific information. In contrast, replacing the complementary views with highly correlated solution prefixes reduces average performance by 1.8\%. This result highlights that \method{} benefits most from genuinely complementary privileged views: highly redundant views provide less additional information and therefore reduce the benefit of multi-view aggregation.

\begin{table}[t]
\centering
\small
\caption{Robustness of \method{} to a misleading privileged view and to highly correlated views on Qwen3-4B.}
\label{tab:view_robustness}
\begin{tabular}{lcccc}
\toprule
\textbf{Setting} & \textbf{AIME24} & \textbf{AIME25} & \textbf{HMMT25} & \textbf{Avg.} \\
\midrule
\method{} (misleading answer view)
& 75.4 & 67.9 & \textbf{44.2} & 62.5 \\
\method{} (90\%/95\%/100\% views)
& 74.6 & 65.4 & 43.3 & 61.1 \\
\method{} (complementary views)
& \textbf{76.2} & \textbf{68.3} & \textbf{44.2} & \textbf{62.9} \\
\bottomrule
\end{tabular}
\end{table}

\subsection{\method{} Analysis}
\paragraph{Gate Analysis.} We analyze how often \method{} uses the residual pathway during training. Specifically, we report the percentage of token positions for which the gate is open, indicating that the method adds residual support from the arithmetic target beyond the geometric consensus signal. As shown in~\cref{tab:gate_open}, the gate-open rate varies across models but remains consistently between 21\% and 27\%. This suggests that \method{} does not simply collapse to either the consensus or arithmetic target: it primarily relies on stable cross-view consensus, while selectively incorporating residual information for tokens where the multi-view signal is reliable.

\begin{figure}[t]
    \centering
    \includegraphics[width=1.0\linewidth]{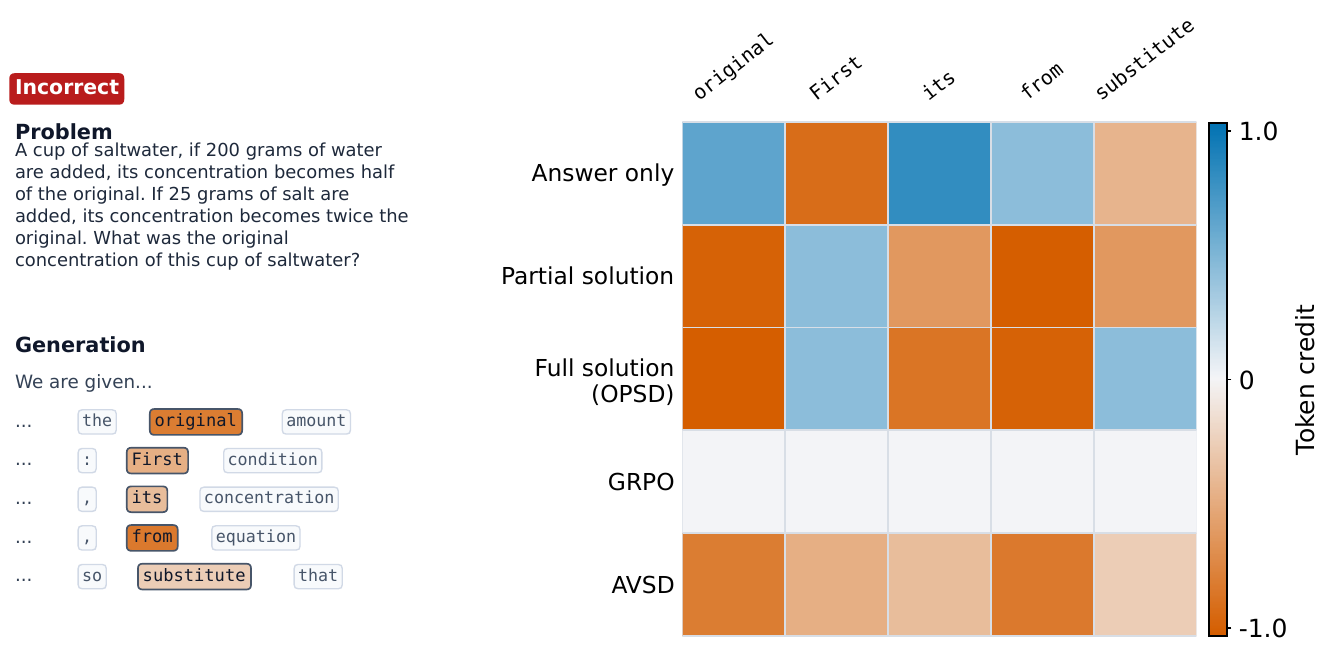}
    \caption{The full problem with generation trace for~\cref{fig:heatmap}.}
    \label{fig:credit}
\end{figure}

\begin{figure*}[t]
    \centering

    \begin{minipage}[t]{0.5\linewidth}
        \centering
        \small
        \begin{tabular}{lc}
        \toprule
        \textbf{Model} & \textbf{Gate Open (\%)} \\
        \midrule
        Qwen3-4B & 24.6 \\
        Qwen3-8B & 21.1 \\
        DeepSeek-R1-Distill-Qwen-7B & 26.6 \\
        \bottomrule
        \end{tabular}
        \captionof{table}{Gate-open rate during training. We report the percentage of token positions for which the residual gate is open.}
        \label{tab:gate_open}
    \end{minipage}
    \hfill
    \begin{minipage}[t]{0.42\linewidth}
        \centering
        \includegraphics[width=\linewidth]{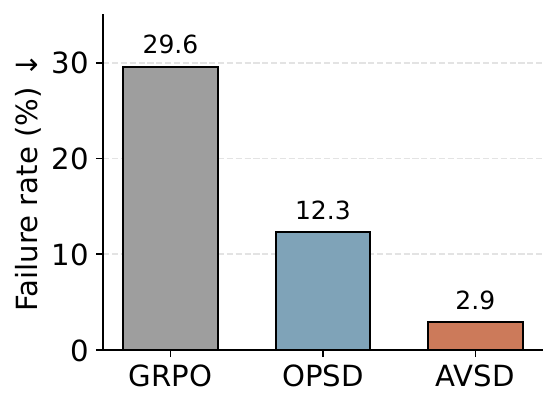}
        \caption{Quantitative reliability of token-level learning signals. For OPSD and \method{}, we report the percentage of top generated tokens with the largest absolute credit magnitude whose assigned credit has the incorrect sign. For GRPO, we report the percentage of cases with no learning signal, which occurs when no sampled rollout is correct. Lower is better.}
        \label{fig:sign}
    \end{minipage}

\end{figure*}

\paragraph{Token-Level Credit Allocation.} We provide supplementary token-level credit results for the main paper in~\cref{fig:credit} and~\cref{fig:sign}. \cref{fig:credit} expands the qualitative example from~\cref{fig:heatmap} by showing the full problem, the generated trace, and the token credits assigned by each method, showing the benefits of \method{} over baselines. \cref{fig:sign} quantifies the reliability of the learning signal: GRPO provides no token-level signal in 29.6\% of incorrect-rollout cases, OPSD assigns incorrect-sign credit to 12.3\% of high-impact tokens, while \method{} reduces this rate to 2.9\%.

\subsection{Additional Baselines and Results}
\paragraph{Multi-Seed Results.}
We further evaluate \method{} and OPSD, the strongest baseline in the Qwen3-4B setting, using five independent training seeds while keeping the training data, hyperparameters, and evaluation configuration fixed. \cref{tab:multiseed} shows that \method{} consistently outperforms OPSD across all three benchmarks, with mean improvements of 2.84, 2.54, and 1.44\% on AIME24, AIME25, and HMMT25, respectively. Moreover, \method{} outperforms OPSD in every paired seed across all three benchmarks, indicating that its gains are consistent across training runs rather than arising from a favorable single seed.

\begin{table}[t]
\centering
\small
\caption{Multi-seed Avg@8 results on Qwen3-4B over five independent training seeds. We report mean $\pm$ standard deviation.}
\label{tab:multiseed}
\begin{tabular}{lccc}
\toprule
\textbf{Method} & \textbf{AIME24} & \textbf{AIME25} & \textbf{HMMT25} \\
\midrule
OPSD
& $72.92 \pm 0.66$
& $65.50 \pm 0.56$
& $42.76 \pm 0.76$ \\
\textbf{\method{} (Ours)}
& $\mathbf{75.76 \pm 0.57}$
& $\mathbf{68.04 \pm 0.58}$
& $\mathbf{44.20 \pm 0.49}$ \\
\bottomrule
\end{tabular}
\end{table}

\paragraph{Confidence-based View-selection Baseline.}
We additionally compare against a parameter-free confidence-based view-selection baseline. At each generated token $y_t$, this baseline selects the view-conditioned teacher assigning the highest probability to that token,
\[
m_t^\star = \arg\max_m q_t^{(m)}(y_t),
\]
and performs standard single-view reverse-KL distillation using the selected teacher $q_t^{(m_t^\star)}$. \method{} outperforms confidence-based selection by 2.7\% on average across AIME24, AIME25, HMMT25. This suggests that simply selecting the teacher assigning the highest probability to each generated token is insufficient: high confidence may reflect view-specific privileged information or overconfidence rather than a reliable learning signal. \method{} instead preserves cross-view consensus and only incorporates view-specific support when it is aligned with and proportionate to the consensus.

\section{Examples of Multi-View Privileged Information} \label{sec:app_view_examples}

\begin{tcolorbox}[
    colback=gray!3,
    colframe=gray!45,
    title={Example of Multi-View Privileged Information in Math},
    fonttitle=\bfseries,
    breakable
]
\textbf{Question.} 		
Real numbers \(a, b, c, d\) are such that \(a + b = \frac{9}{c - d}\) and \(c + d = \frac{25}{a - b}\). What is the minimum value that the quantity \(a^2 + b^2 + c^2 + d^2\) can take?

\vspace{0.5em}
\textbf{Full Solution View}
Given the equations involving real numbers \(a, b, c,\) and \(d\):

\[ a + b = \frac{9}{c - d} \]
\[ c + d = \frac{25}{a - b} \]

We aim to find the minimum value of \(a^2 + b^2 + c^2 + d^2\).

1. **Multiply both given equations by their respective denominators**:

\[
(a + b)(c - d) = 9
\]

\[
(c + d)(a - b) = 25
\]

2. **Expand and rearrange the products**:

\[
ac - ad + bc - bd = 9
\]

\[
ac - bc + ad - bd = 25
\]

3. **Add the two equations to eliminate terms**:

\[
2(ac - bd) = 34 \implies ac - bd = 17
\]

4. **Establish a lower bound for \(a^2 + b^2 + c^2 + d^2\)**:

Since \(a^2 + c^2 \geq 2ac\) (which follows from the AM-GM inequality \( (a - c)^2 \geq 0 \)) and
\(b^2 + d^2 \geq -2bd\) (which follows from \( (b + d)^2 \geq 0 \)):

\[
a^2 + c^2 \geq 2ac
\]

\[
b^2 + d^2 \geq -2bd
\]

Adding these inequalities yields:

\[
a^2 + b^2 + c^2 + d^2 \geq 2(ac - bd)
\]

Given \( ac - bd = 17 \):

\[
a^2 + b^2 + c^2 + d^2 \geq 2 \times 17 = 34
\]

5. **Check if the equality can be achieved**:

Equality in the inequalities $a^2 + c^2 \geq 2ac$ and $b^2 + d^2 \geq -2bd$ is achieved when $a = c$ and $b = -d$.

Substitute $a = c$ and $b = -d$ into the original equations. The first equation becomes:
\[
a + b = \frac{9}{c - d} \implies a - d = \frac{9}{a - d} \implies (a - d)^2 = 9.
\]
The second equation becomes:
\[
c + d = \frac{25}{a - b} \implies a + d = \frac{25}{a + d} \implies (a + d)^2 = 25.
\]
Taking positive roots gives $a - d = 3$ and $a + d = 5$, so $a = 4$ and $d = 1$. Hence $c = a = 4$ and $b = -d = -1$. Substituting back:
\[
a^2 + b^2 + c^2 + d^2 = 4^2 + (-1)^2 + 4^2 + 1^2 = 16 + 1 + 16 + 1 = 34.
\]
This confirms the minimum value is achievable.

Conclusion
\[
\boxed{34}
\]

\vspace{0.5em}
\textbf{Partial Solution View}
Given the equations involving real numbers \(a, b, c,\) and \(d\):

\[ a + b = \frac{9}{c - d} \]
\[ c + d = \frac{25}{a - b} \]

We aim to find the minimum value of \(a^2 + b^2 + c^2 + d^2\).

1. **Multiply both given equations by their respective denominators**:

\[
(a + b)(c - d) = 9
\]

\[
(c + d)(a - b) = 25
\]

2. **Expand and rearrange the products**:

\[
ac - ad + bc - bd = 9
\]

\[
ac - bc + ad - bd = 25
\]

\vspace{0.5em}
\textbf{Final Answer View}
\[
\boxed{34}
\]
\end{tcolorbox}

\begin{tcolorbox}[
    colback=gray!3,
    colframe=gray!45,
    title={Example of Multi-View Privileged Information in Math},
    fonttitle=\bfseries,
    breakable
]
\textbf{Question.} 		
In a company, employees have a combined monthly salary of \$10,000. A kind manager proposes to triple the salaries of those earning up to \$500, and to increase the salaries of others by \$1,000, resulting in a total salary of \$24,000. A strict manager proposes to reduce the salaries of those earning more than \$500 to \$500, while leaving others' salaries unchanged. What will the total salary be in this case?

\vspace{0.5em}
\textbf{Full Solution View}

1. Let us denote by $x$ the total number of employees whose monthly salary is less than or equal to 500 dollars.

2. Let $y$ be the total number of employees whose monthly salary is more than 500 dollars.

3. The firm's total monthly salary expenditure before any changes is 10,000 dollars. This can be written as:
\[
500x + \sum_{i=1}^{y} s_i = 10,000
\]
where $s_i$ represents the individual salaries of the employees earning more than 500 dollars.
4. The kind manager proposes the following changes:
- Tripling the salaries of employees earning 500 dollars or less.
- Increasing the salaries of employees earning more than 500 dollars by 1000 dollars each.

The new total salary expenditure proposed by the kind manager is 24,000 dollars. This can be expressed as:
\[
3 \cdot 500x + \sum_{i=1}^{y} (s_i + 1000) = 24000
\]
5. Simplifying the equation:
\[
1500x + \sum_{i=1}^{y} s_i + 1000y = 24000
\]
6. Since we know that $\sum_{i=1}^{y} s_i = 10000 - 500x$ from the original total monthly salary equation, substituting this we get:
\[
1500x + (10000 - 500x) + 1000y = 24000
\]
7. Simplifying further, we obtain:
\[
1000x + 1000y = 14000 \implies x + y = 14
\]
8. The cruel manager proposes the following changes:
- Reducing the salaries of employees earning more than 500 dollars to 500 dollars each.
- Keeping the salaries of employees earning 500 dollars or less unchanged.

The new total salary can be calculated as:
\[
500x + 500y = 500(x + y)
\]
9. From step 7, we know that $x + y = 14$, so substituting this into the equation for the total salary proposed by the cruel manager:
\[
500 \cdot 14 = 7000
\]
10. Conclusion:
The new total salary expenditure after the cruel manager’s proposed changes will be:
\[
\boxed{7000}
\]

\vspace{0.5em}
\textbf{Partial Solution View}

1. Let us denote by $x$ the total number of employees whose monthly salary is less than or equal to 500 dollars.

2. Let $y$ be the total number of employees whose monthly salary is more than 500 dollars.

3. The firm's total monthly salary expenditure before any changes is 10,000 dollars. This can be written as:
\[
500x + \sum_{i=1}^{y} s_i = 10,000
\]
where $s_i$ represents the individual salaries of the employees earning more than 500 dollars.

\vspace{0.5em}
\textbf{Final Answer View}
\[
\boxed{7000}
\]
\end{tcolorbox}

\begin{tcolorbox}[
    colback=teal!3,
    colframe=teal!55!black,
    title={Example of Multi-View Privileged Information in Code},
    fonttitle=\bfseries,
    breakable
]
\textbf{Question.} \\
Given \(n\) sticks with lengths \(a_i\), find the maximum number of regular equal-sided polygons
you can construct. Each side uses one stick, and no stick can be reused.

\vspace{0.5em}
\textbf{Reference Solution View}

\begin{lstlisting}[basicstyle=\ttfamily\small, breaklines=true]
import sys
from collections import Counter

t = int(sys.stdin.readline())
for _ in range(t):
    n = int(sys.stdin.readline())
    a = list(map(int, sys.stdin.readline().split()))
    cnt = Counter(a)
    total = 0
    for v in cnt.values():
        total += v // 3
    print(total)
\end{lstlisting}

\vspace{0.5em}
\textbf{Hint View}

Greedily form triangles from sticks of the same length. For \(S\) sticks of one length,
you can create \(\lfloor S / 3 \rfloor\) polygons. Sum this over all stick lengths.

\vspace{0.5em}
\textbf{Feedback View}

Wrong student rollout:

\begin{lstlisting}[basicstyle=\ttfamily\small, breaklines=true]
import sys
input=sys.stdin.readline
t=int(input())
for _ in range(t):
    n=int(input())
    a=list(map(int,input().split()))
    print(0)
\end{lstlisting}

Feedback: The student's solution failed on a test case.

Failing input:
\begin{lstlisting}[basicstyle=\ttfamily\small, breaklines=true]
4
1
1
2
1 1
6
2 2 3 3 3 3
9
4 2 2 2 2 4 2 4 4
\end{lstlisting}

Expected output:
\begin{lstlisting}[basicstyle=\ttfamily\small, breaklines=true]
0
0
1
2
\end{lstlisting}

Student output:
\begin{lstlisting}[basicstyle=\ttfamily\small, breaklines=true]
0
0
0
0
\end{lstlisting}

\end{tcolorbox}

\begin{tcolorbox}[
    colback=teal!3,
    colframe=teal!55!black,
    title={Example of Multi-View Privileged Information in Code},
    fonttitle=\bfseries,
    breakable
]
\textbf{Question.} \\
\(S(x)\) is the sum of digits of \(x\). Count integers \(x\) with \(1 \le x \le n\) such that
\[
S(x+1) < S(x).
\]

\vspace{0.5em}
\textbf{Reference Solution View.}

\begin{lstlisting}[basicstyle=\ttfamily\small, breaklines=true]
t = int(input())
for _ in range(t):
    n = int(input())
    print((n + 1) // 10)
\end{lstlisting}

\vspace{0.5em}
\textbf{Hint View.}

A number is interesting exactly when it ends in \(9\). Count numbers \(\le n\) whose last
digit is \(9\), which is \(\lfloor (n+1)/10 \rfloor\).

\vspace{0.5em}
\textbf{Feedback View.}

Wrong student rollout:

\begin{lstlisting}[basicstyle=\ttfamily\small, breaklines=true]
t=int(input())
for _ in range(t):
    n=int(input())
    print(n//10)
\end{lstlisting}

Feedback: The student's solution failed on a test case.

Failing input:
\begin{lstlisting}[basicstyle=\ttfamily\small, breaklines=true]
5
1
9
10
34
880055535
\end{lstlisting}

Expected output:
\begin{lstlisting}[basicstyle=\ttfamily\small, breaklines=true]
0
1
1
3
88005553
\end{lstlisting}

Student output:
\begin{lstlisting}[basicstyle=\ttfamily\small, breaklines=true]
0
0
1
3
88005553
\end{lstlisting}

\end{tcolorbox}

\section{Prompts} \label{sec:app_prompts}
\begin{tcolorbox}[
    colback=purple!3,
    colframe=purple!55!black,
    title={Teacher Prompt},
    fonttitle=\bfseries,
    breakable
]
\texttt{Problem: \{problem\}}

\vspace{0.5em}
\texttt{Here is training-time reference information for this problem. It may contain a final answer, a hint, a partial solution, or a full solution to the problem.}

\vspace{0.5em}
\texttt{=== Privileged Reference (\{view\_type\}) Begin ===}

\texttt{\{privileged information\}}

\texttt{=== Privileged Reference (\{view\_type\}) End ===}

\vspace{0.5em}
\texttt{Use the reference only as hidden guidance. Use it only to understand, verify, or correct your solution.}

\texttt{Do not mention, quote, copy, cite, or refer to the privileged reference or its existence.}

\texttt{Now solve the original problem independently using your own reasoning.}

\texttt{Please reason step by step, and put your final answer within \textbackslash boxed\{\}.}
\end{tcolorbox}

\vspace{0.75em}

\begin{tcolorbox}[
    colback=orange!4,
    colframe=orange!65!black,
    title={Student Prompt},
    fonttitle=\bfseries,
    breakable
]
\texttt{Problem: \{problem\}}

\vspace{0.5em}
\texttt{Please reason step by step, and put your final answer within \textbackslash boxed\{\}.}
\end{tcolorbox}

\end{document}